\pdfoutput=1
\PassOptionsToPackage{table}{xcolor}
\documentclass[11pt]{article}

\usepackage{acl}
\usepackage[table]{xcolor}
\usepackage{newtxtext,newtxmath}

\usepackage{latexsym}
\usepackage{amsmath}
\usepackage{booktabs}
\usepackage{multirow}
\usepackage{algorithm}
\usepackage{algpseudocode}
\usepackage{tikz}
\usepackage{makecell}

\newcommand*\circled[1]{\tikz[baseline=(char.base)]{
            \node[shape=circle,fill=gray!30,inner sep=2pt] (char) {#1};}}

\usepackage[T1]{fontenc}

\usepackage[utf8]{inputenc}

\usepackage{microtype}

\usepackage{inconsolata}

\usepackage{graphicx}

\title{
Large Language Models are Easily Confused:\\
A Quantitative Metric, Security Implications and Typological Analysis
}

\author{
 \textbf{Yiyi Chen\textsuperscript{1}},
 \textbf{QiongXiu Li\textsuperscript{2}},
 \textbf{Russa Biswas\textsuperscript{1}},
 \textbf{Johannes Bjerva\textsuperscript{1}}
\\
 \textsuperscript{1}Department of Computer Science,
 \textsuperscript{2}Department of Electronic Systems\\
 Aalborg University, Copenhagen, Denmark
\\
 \small{
   \texttt{\{yiyic, rubi, jbjerva\}@cs.aau.dk, qili@es.aau.dk}
 }
}

\begin{document}
\maketitle
\raggedbottom

\begin{abstract}
\textit{Language Confusion} is a phenomenon where Large Language Models (LLMs) generate text that is \textbf{neither} in the desired language, \textbf{nor} in a contextually appropriate language. 
This phenomenon presents a critical challenge in text generation by LLMs, often appearing as erratic and unpredictable behavior. 
We hypothesize that there are linguistic regularities to this inherent vulnerability in LLMs and shed light on patterns of language confusion across LLMs.
We introduce a novel metric, \textit{Language Confusion Entropy}, designed to directly measure and quantify this confusion, based on language distributions informed by linguistic typology and lexical variation. 
Comprehensive comparisons with the Language Confusion Benchmark ~\citep{marchisio2024understanding} confirm the effectiveness of our metric, revealing patterns of language confusion across LLMs.
We further link language confusion to LLM security and find patterns in the case of multilingual embedding inversion attacks.
Our analysis demonstrates that linguistic typology offers theoretically grounded interpretation, and valuable insights into leveraging language similarities as a prior for LLM alignment and security.\footnote{The language graphs for language similarities and code are publicly available~\url{https://github.com/siebeniris/QuantifyingLanguageConfusion/}}
\end{abstract}

\section{Introduction}
Multilingual Large Language Models (LLMs) revolutionized Natural Language Processing (NLP), offering crosslinguality in various applications, including translation~\cite{zhu2024multilingual}, text generation~\cite{chen2022mtg}, and information retrieval~\cite{guo2024steering}. Besides the challenges faced by LLMs such as \textit{bias and fairness}~\cite{talat2022you}, \textit{hallucinations}~\cite{augenstein2024factuality}, multilingual LLMs are more vulnerable to \textit{adversarial and inversion attacks} than monolingual LLMs~\cite{song2024multilingual,chen2024oddsovercomingtypologyscript,chen-etal-2024-text}. 

Multilingual LLMs are trained on data in a diverse range of languages to represent the intricacies of multiple languages within a single model. 
However, this often results in inconsistencies in comprehension and response, leading to \textit{language confusion} -- instances where LLMs generate text that is \textit{neither} in the desired language \textit{nor} in a contextually appropriate one. 
For example, when an LLM is queried/prompted in Arabic, it may respond in text that is either partially or entirely in languages other than Arabic, e.g., English.

\citet{marchisio2024understanding} propose metrics to measure the percentage of model responses containing no undesired languages at both line and word levels but fail to capture nuances within language distributions. 
It is observed that language confusion tends to occur when the model's distribution over the next tokens is flat.
We hypothesize that language confusion in LLMs is not merely a performance limitation but an inherent vulnerability, partly due to imbalanced pre-training multilingual data sources, which are amplified with increasing numbers of languages and can be analyzed through language similarities derived from linguistic typology and other resources.

To thoroughly investigate language confusion as a phenomenon and its role within LLMs, we introduce the following research questions:

\noindent\textbf{RQ1}: What measurable patterns characterize language confusion in LLMs, and how can these patterns be quantified effectively?

\noindent\textbf{RQ2}: How do language similarities influence language confusion, and how can this knowledge be applied to enhance LLM alignment and security?

To this end, we propose a novel metric called \textit{Language Confusion Entropy}, which provides a quantifiable measure of uncertainty and facilitates the detection of when an LLM is confused. Building on observations by~\citet{marchisio2024understanding} that uniformity of the distribution indicates higher uncertainty, \textit{Language Confusion Entropy}
re-weights language distributions by emphasizing long-tail distributions, effectively capturing language confusion in multilingual LLM generation tasks. 
Furthermore, we demonstrate that this metric uncovers patterns of language confusion during both the training and evaluation phases of multilingual inversion attacks~\citep{chen2024oddsovercomingtypologyscript}.

In addition, we construct language graphs based on language similarities derived from linguistic resources to analyze language confusion, revealing a strong correlation between language confusion and semantic similarities between languages.
Our analysis shows that low-resource languages exhibit less confusion while training across diverse scripts and language families mitigates language confusion more effectively than training within the same script or language family in inversion attacks. 
These findings indicate that leveraging language similarities grounded in linguistic resources could serve as a valuable prior for enhancing LLM alignment and security. Our main contributions are as follows:

\noindent\textbf{1)} We propose a novel metric \textit{Language Confusion Entropy} to measure language confusion in LLMs considering the nuances of language distributions. To the best of our knowledge, we are the first to quantify language confusion probabilistically.

\noindent\textbf{2)} Using language graphs, we demonstrate that linguistic typology provides a foundational tool for analyzing language confusion.

\noindent\textbf{3)}  We propose a modified KL-Divergence algorithm to determine the correlation between language similarities (as defined by language graphs) and language confusion in LLMs.

\noindent\textbf{4)} We conduct extensive analysis revealing statistically significant patterns of language confusion, providing new insights for LLM security research.

\section{Related Works}
\paragraph{Language Confusion}
This phenomenon observed in  NLP, is often described as ``off-target translation''~\citep{chen-etal-2023-target, sennrich-etal-2024-mitigating} or ``accidental translation''~\citep{zhang-etal-2020-improving, xue2020mt5}, or as ``source language hallucinations'' in zero-shot transfer scenarios~\citep{vu_overcoming_2022, li-murray-2023-zero, pfeiffer-etal-2023-mmt5, chirkova-nikoulina-2024-key}.
\textit{Language confusion}, a term coined by~\citet{marchisio2024understanding} occurs when the LLMs' outputs are generated \textit{erroneously} in languages different from the desired (target) languages and identified as `surprising limitation' diminishing LLM utility for non-English languages, 
indicating the \textit{unpredictable nature}. 


The phenomenon of \textit{language confusion} has not only been pervasive in LLMs, 
but also in tasks pertinent to LLM security, such as multilingual inversion attacks~\citep{chen-etal-2024-text, chen2024oddsovercomingtypologyscript}. Furthermore, \citet{chen2024oddsovercomingtypologyscript} observes language confusion across 20 languages from diverse scripts and language families in multilingual embedding inversion. They analyzed the pattern of confusion using basic typological features between train and eval languages with regression analysis, in comparison, the proposed \textit{Language Confusion Entropy} provides a more interpretable analysis.

\paragraph{Multilingual LLM Safety and Security}
~\citet{yong_low-resource_2024} exposes vulnerabilities of AI safety mechanism by jailbreaking GPT-4's safeguard through translating unsafe English inputs into low-resource languages.~\citet{deng_multilingual_2024} impose unintentional and intentional jailbreak on multilingual LLMs, using multilingual prompts. 
It is observed that low-resource languages are more vulnerable, making them the weakest links in AI security.

Backdoor attacks on multilingual machine translation pose significant threats, as injecting poisoned data into low-resource language pairs can achieve a high attack success rate (ASR) in high-resource language pairs~\citep{wang_backdoor_2024}. 
Poisoning instruction-tuning data for one or two languages can affect other languages, surpassing 99\% ASR in the cross-lingual setting in prominent LLMs resisting current defenses~\citep{he_tuba_2024}.

Multilingual textual embedding inversion attacks pose additional risks, as any encoder can be attacked to reconstruct original texts. Traditional defenses for monolingual LLMs are ineffective for multilingual LLMs~\citep{chen-etal-2024-text, chen2024oddsovercomingtypologyscript}. Moreover,~\citet{song2024multilingual} generates language blending for adversarial attacks, necessitating systematic analysis of language similarity and language confusion for targeted defenses.

\paragraph{Linguistic Typology and Language Similarities}
Previous research on multilingual effects on linguistic level uses three approaches:
(i) phylogenetic variation, (ii) linguistic typological variation, and (iii) embedded and data-driven language variation.

While genealogical relations
are intuitive, the correlations between language similarity and genealogical relations are often spurious~\citep{rama-kolachina-2012-good}.
\citet{ploeger2024typological} challenge this approach highlighting its negative impact on downstream NLP tasks.

Linguistic typology offers a theoretically grounded approach to measuring similarity between languages \citep{kashyap2019language}.
Languages can be categorized based on various features --- e.g., whether they use a Subject-Verb word order (SV) or the opposite (VS).
Such information has been manually annotated in linguistic databases, including WALS~\citep{haspelmath2008typological}, ASJP ~\citep{wichmann2012automated}, and Grambank ~\citep{skirgaard2023grambank}, and work in NLP has contributed with automatic prediction of such features \citep{malaviya-etal-2017-learning,bjerva-etal-2019-probabilistic,bjerva-2024-role}.
Recent work has also explored lexically driven measures, 
showing that multilingual LLMs often rely on lexical overlap~\citep{pires-etal-2019-multilingual}. 
Such work spans from using synonymy-relations as in WordNet \citep{fellbaum2010wordnet}, to multilingual relations in BabelNet~\citep{navigli2010babelnet}, and more complex colexification patterns in CLICS$^3$~\citep{rzymski2020database}.
Crosslingual colexification patterns refer to the phenomenon whereby different meanings are captured by the same lexica across languages~\citep{franccois2008semantic}, implicating shared cognitive or cultural associations~\citep{karjus2021conceptual,di2021colexification,chen-bjerva-2023-colexifications}.

Language embeddings encode language characteristics and can be derived from data-driven methods or linguistic typological databases.~\citet{ostling-tiedemann-2017-continuous} trained a character-level LSTM language model 
on translated Bible texts from 990 languages, showing their ability to reconstruct language genealogies. Additionally, embeddings can be generated from typological data sources like WALS, Grambank, and ASJP. \citet{chen-etal-2023-colex2lang} created embeddings using lexical data based on colexification patterns in CLICS3 and BabelNet.

We hypothesize that language confusion is a phenomenon largely driven by lexical variation, similar to the patterns observed by \citet{pires-etal-2019-multilingual}.
We investigate this by building our analysis on this body of work leveraging computational typology.

\section{Explainable Language Confusion}

To investigate the phenomenon of language confusion, we use the datasets i) \textbf{Language Confusion Benchmark (LCB)}~\citep{marchisio2024understanding} and 
ii) \textbf{Multilingual Textual Embedding Inversion (MTEI)}~\citep{chen2024oddsovercomingtypologyscript} (see Appendix for task details and Table~\ref{tab:examples_mtei} for processed datasets sample).

\begin{figure}[h!]
    \includegraphics[width=\linewidth]{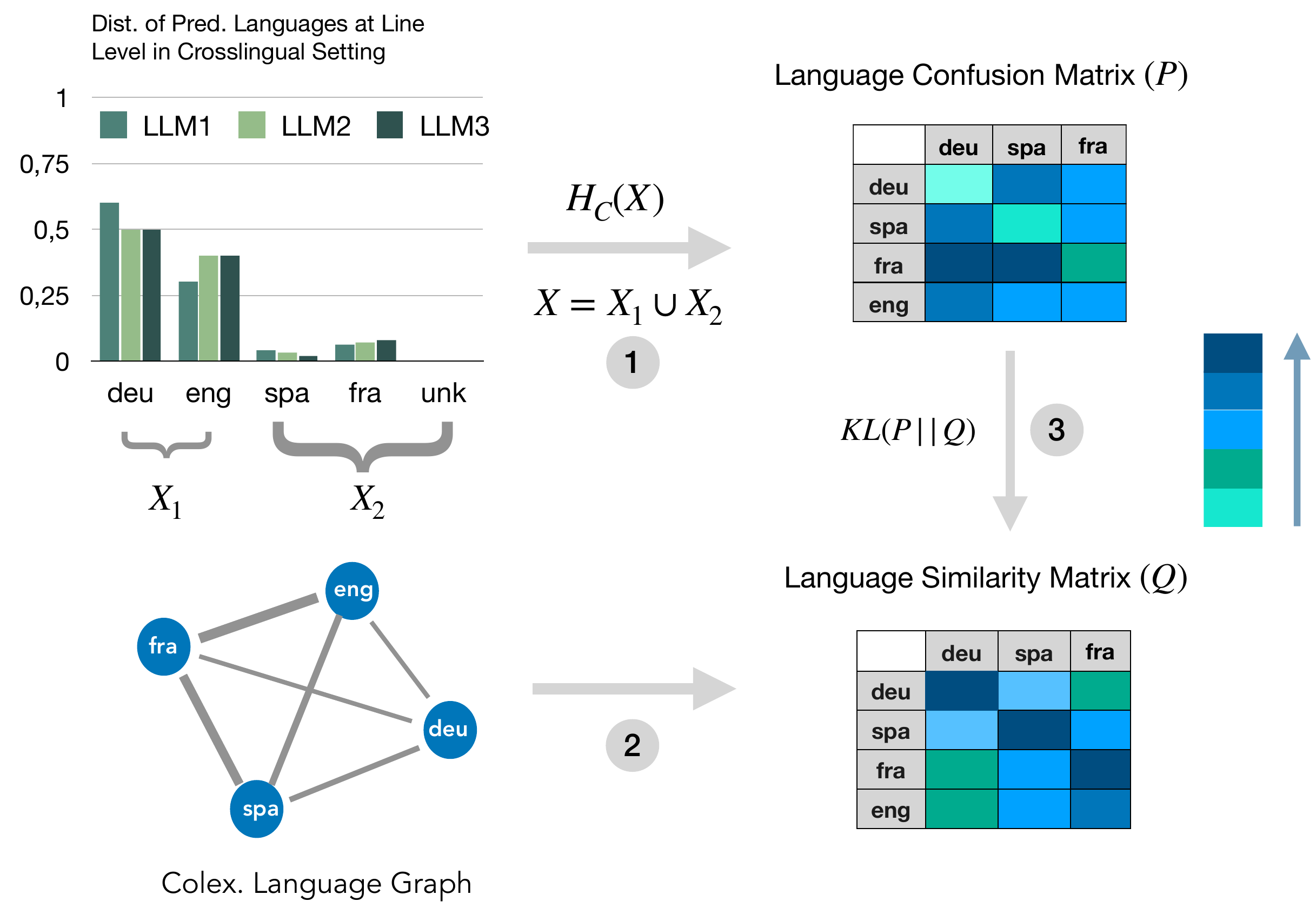}
    \caption{The use of proposed metric to quantify language confusion and its correlation with language similarity through KL divergence. }
    \label{fig:method}
\end{figure}

\subsection{Generation Settings} Consider a LLM trained or prompted with a set of $n \in \mathbb{N}$ source languages $L_s = \{l_1, \cdots, l_n\}$, and $l_t$ is the target language. We probe language confusion for both LLM instruction and textual embedding inversion attacks.

\paragraph{Monolingual Generation}
\textbf{LCB}: The model is queried in language $l_t$, and the response is expected in $l_t$.
\textbf{MTEI}: The inversion model, trained on $l_t$, inverts embeddings in $l_t$. Here, $l_t \in L_s$, i.e., the evaluated language is part of the training languages.

\paragraph{Crosslingual Generation} 
\textbf{LCB}: The model is instructed in language $l_s$ to provide a response in $l_t$, where $l_t\neq l_s$.
\textbf{MTEI}: The inversion model, trained on $L_s$, inverts embeddings in $l_t$. 
In this setting, $l_t \notin L_s$, meaning the evaluated language differs from the training languages. 

\subsection{Quantifying Language Confusion}
The phenomenon of language confusion is particularly prominent in crosslingual generation settings.
For instance in Fig.~\ref{fig:method}~\circled{1}, when an LLM is prompted in English and expected to generate a response in German ($X_1$), the output may unexpectedly have a mix of other languages, such as Spanish and French ($X_2$). Ideally, the LLM should focus on the expected languages; thus, the model exhibits greater confusion if its output distribution assigns high probabilities to unexpected languages. To quantify this, we propose \textit{Language Confusion Entropy} ($H_{\mathbf{C}}$), defined as follows:
\begin{equation} 
\begin{aligned} 
H_{\mathbf{C}}(X) = & -\sum_{x \in X_1} (1 - p(x)) \log(p(x)) \\ & - \sum_{x \in X_2} p(x) \log p(x), \end{aligned}
\label{eq:confusion_entropy} 
\end{equation} 
where $X_1$ denotes the expected language set and $X_2$ the unexpected language set, $X_1 \cup X_2= X, ~X_1 \cap X_2 =\emptyset$, $p(x)$ denotes the probability and $\sum_{x\in X} p(x) = 1$. 
Ideally, an unconfused model would satisfy $\sum_{x\in X_1} p(x) = 1$ and $\sum_{x\in X_2} p(x) = 0$.

In practical applications, for the \textbf{LCB} dataset, $X_1$ includes both the instruction and response languages, while for the \textbf{MTEI} dataset, $X_1$ encompasses the training and evaluation languages.

\subsection{Language Identification}
Language confusion can occur at both the word level and the line level.
To measure it, we use existing language identification (LID) tools. 
Initially, we employ Lingua to detect languages as it offers the highest accuracy among existing LID tools, particularly at word level.\footnote{\url{https://github.com/pemistahl/lingua-py}}
However, since Lingua supports only 75 languages, we supplement it with fastText~\citep{joulin2016bag}
(which supports 176 languages) for languages unidentified by Lingua.


\paragraph{Line-level Detection}
We split the generated output into lines by newline characters and detect the language of each line. 

\paragraph{Word-level Detection}
To detect languages more accurately at the word level, we first tokenize the text at the line level using language-specific tokenizers such as jieba~\citep{jieba_chinese}
(Chinese), 
Hebrew Tokenizer~\citep{hebrew_tokenizer}
Kiwipiepy~\citep{kiwi-korean} (Korean),
fugashi~\citep{mccann-2020-fugashi} (Japanese), 
and NLTK Word Tokenizer
~\citep{bird-loper-2004-nltk} for other languages.
We then identify the language of each tokenized word. 

Finally, we compile the detected languages into distributions at both word and line levels for further analysis. Examples of the pre-processed data are presented in Table~\ref{tab:examples_mtei}. Language confusion matrices are then constructed by applying language confusion entropy to these distributions, and the results are aggregated per language (ref. Fig.~\ref{fig:method}~\circled{1}).

\begin{algorithm}
\small
\caption{KL Divergence for Language Confusion vs. Language Similarity Matrices}\label{alg:kl_divergence_matrices}
\begin{algorithmic}[1]
\Require Matrices \( \text{M1} \in \mathbf{R}^{n \times m} \) and \( \text{M2} \in \mathbf{R}^{n \times m} \), where M1 represent language-to-language confusion scores, and M2 represent language-to-language similarity scores.
\Ensure Mean KL divergence across all columns.
\State Initialize a list \(\text{Total\_KL\_Divergence} \gets [ ]\).
\For{\textbf{each} column index \( j \) from \( 1 \) to \( m \)}
    \State \(\text{M1\_col} \gets \text{M1}[:, j]\) \Comment{Confusion scores for language \( j \) in matrix M1}
    \State \(\text{M2\_col} \gets \text{M2}[:, j]\) \Comment{Similarity scores for language \( j \) in matrix M2}
    \State \textbf{Step 1: Exclude zeros from M1\_col}
    \State \(\text{nonzero\_indices} \gets \text{M1\_col} \neq 0\)
    \State \( P \gets \text{M1\_col}[\text{nonzero\_indices}]\)
    \State \( Q \gets \text{M2\_col}[\text{nonzero\_indices}]\)
    \State \textbf{Step 2: Normalize the distributions}
    \State \( P \gets P / \sum P \) 
    \State \( Q \gets Q / \sum Q \)
    \State \textbf{Avoid division by zero or log issues}
    \State \(\epsilon \gets 10^{-10}\)
    \State \( P \gets P + \epsilon \)
    \State \( Q \gets Q + \epsilon \)
    \State \textbf{Step 3: Calculate KL divergence}
    \State \(\text{KL\_Div} \gets \sum P \cdot \log\left(\frac{P}{Q}\right)\) \Comment{KL divergence for \( j \)}
    \State \text{Append \(\text{KL\_Div}\) to \(\text{Total\_KL\_Divergence}\)}
\EndFor
\State \Return \(\text{Average}(\text{Total\_KL\_Divergence})\)
\end{algorithmic}
\end{algorithm}

\subsection{Large Language Models}
The evaluated LLMs in~\textbf{LCB} include 
Command R (35B parameters),
Command R+ (104B), 
GPT-3.5 Turbo~\citep{brown2020language},
and GPT-4 Turbo~\citep{achiam2023gpt},
Mistral Large, 
Mistral 8x7B~\citep{jiang2024mixtral},
LLaMA 2 70B Instruct~\citep{touvron2023llama},
and LLaMA 3 70B Instruct, 
while in ~\textbf{MTEI} the inversion models are trained with mT5 (580M)~\citep{xue2020mt5} and multilingual-e5-base (\textsc{me5}) (580M). 
(See Table~\ref{tab:llms} for details of LLMs).

\subsection{Language Graphs}
We construct language graphs from a diverse range of typological features, such as colexification patterns~\citep{rzymski2020database, fellbaum2010wordnet, navigli2010babelnet}, lexicon~\citep{wichmann2012automated}, phonological and morphological-syntactical features~\citep{haspelmath2008typological,skirgaard2023grambank}, as well as from a collection of existing language embeddings trained from NLP tasks incorporating linguistic typology~\citep{ostling-tiedemann-2017-continuous,ostling-kurfali-2023-language,chen-etal-2023-colex2lang}. We then generate language similarity matrices from the language graphs by calculating pairwise similarity using either Jaccard Index or Cosine Similarity (ref. Fig.~\ref{fig:method}~\circled{2}). (See details in Appendix~\ref{sec:language_graphs}).


\subsection{The Role of Language Similarity in Language Confusion}
We compare the language graphs with language confusion matrices to assess how well language confusion aligns with language similarities and to identify specific aspects where they match.
To quantify the divergence between language confusion (denoted as $P$) and language similarity (denoted as $Q$), we employ Kullback-Leibler Divergence~\citep{kullback1951information}, expressed as $KL(P||Q)$. 
$P(x)$ represents the distribution of language confusion entropy of language $x$ relative to other languages, while $Q(x)$ represents the distribution of language similarity of $x$ relative to other languages. 
$KL(P||Q)$ is computed as follows:
\begin{equation}
    \begin{aligned}
   KL(P||Q) = \sum_{x}P(x)log \left(\frac{P(x)}{Q(x)}\right),
    \end{aligned}
    \label{eq:kl_divergence}
\end{equation}
where a lower $KL(P||Q)$ indicates a stronger correspondence between language confusion patterns and underlying language similarities.
(See Algorithm~\ref{alg:kl_divergence_matrices} for detailed steps and Fig~\ref{fig:method}~\circled{3}).

\begin{figure*}[t!]
    \includegraphics[width=0.95\linewidth]{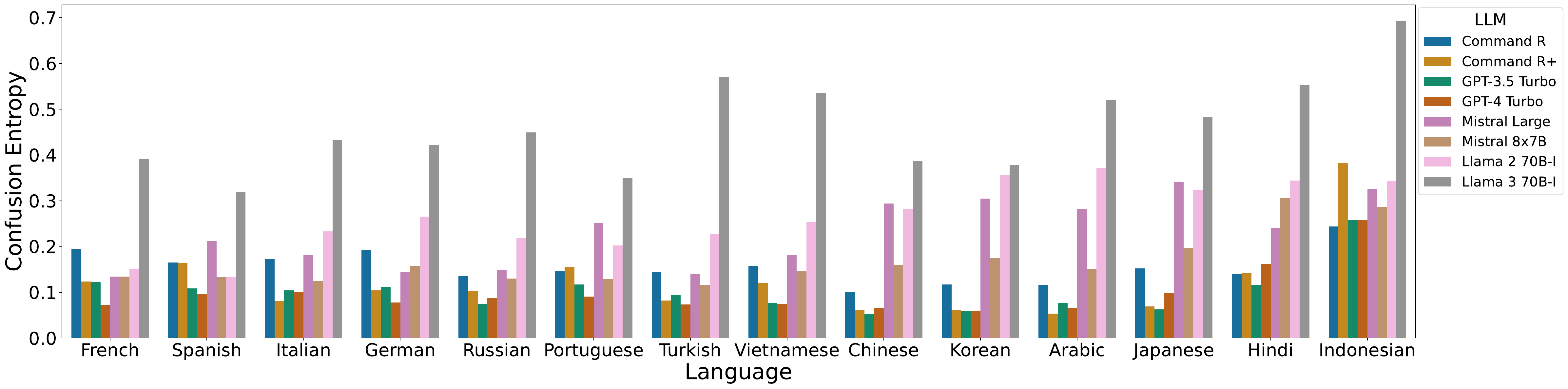}
    \caption{Language confusion for \textbf{LCB} by each language across LLMs for crosslingual setting at Line level. The languages are ordered ascendingly by their language confusion entropy averaged across LLMs.}
    \label{fig:lcb_line_crosslingual_per_lang}
\end{figure*}

\section{Analysis and Results}

\subsection{Language Confusion in LLM Prompting}~\label{sec:lc_llm_prompting}

\paragraph{Language Confusion Entropy vs. Pass Rates}

The binary metrics, \textbf{Pass Rates} at line-level~\textbf{LPR} and word-level~\textbf{WPR} are used to evaluate whether the LLM output contains no error, following~\citet{marchisio2024understanding} (see details in Appendix~\ref{sec:datasets}).
We apply language confusion entropy to \textbf{LCB}, calculating it at both the line-level~\textbf{$H_{\mathbf{C}}$[L]} and word-level~\textbf{$H_{\mathbf{C}}$[W]} across generation settings. 

Compared to \citet{marchisio2024understanding}, our approach detects language confusion across all languages, including at the word level, by using language-specific tokenizers and a more accurate language identification (LID) tool.
We reproduce \textbf{LPR} and \textbf{WPR} (ref. Table~\ref{tab:lcb_lpr_reproduction},~\ref{tab:lcb_wpr_reproduction}) 
and compute~\textbf{$H_{\mathbf{C}}$[L]} and~\textbf{$H_{\mathbf{C}}$[W]} (ref. Table~\ref{tab:lcb_confusion_entropy_llms})  in both crosslingual and monolingual settings, for 14 languages and 8 LLMs, following~\citep{marchisio2024understanding}.

To evaluate the efficacy of language confusion entropy compared to pass rate metrics, we calculate the Spearman correlation\footnote{\url{https://docs.scipy.org/doc/scipy/reference/generated/scipy.stats.spearmanr.html}} coefficients between these metrics across levels and generation settings. 
Overall,~\textbf{$H_{\mathbf{C}}$[L]} shows a strong negative correlation with~\textbf{LPR} across all generation settings.
Moreover, \textbf{$H_{\mathbf{C}}$[W]} - which is based on more detailed language distributions - exhibits a weaker correlation with \textbf{WPR}, 
as \textbf{WPR} only considers English words in non-Latin script languages.
Despite the weaker correlation, it remains statistically significant for all languages, especially crosslingually.

At both the line and word levels, $H_{\mathbf{C}}$ shows a stronger correlation with pass rates in crosslingual settings than in monolingual ones. 
This aligns with the definition of language confusion entropy, which gives more weight to long-tail distributions, a more prominent phenomenon in crosslingual tasks.

\begin{table}[t!]
    \centering
     \resizebox{0.85\linewidth}{!}{ 
    \begin{tabular}{c|c|cccc}
    \toprule 
        & & \textbf{$H_{\mathbf{C}}$[L]} &   \textbf{$H_{\mathbf{C}}$[W]} & \textbf{LPR}   \\
        \midrule
   \multirow{3}{*}{\textbf{All}} &
        \textbf{$H_{\mathbf{C}}$[W]} & $0.51$*** &  & \\
        & \textbf{LPR} & $-0.83$*** &  $-0.3$*** &  \\
        & \textbf{WPR} & $-0.29$**	& $\mathbf{-0.5}$***	 & $0.31$** \\
        \midrule
        \multirow{3}{*}{\textbf{Monolingual}} &
        \textbf{$H_{\mathbf{C}}$[W]} & $0.42$*** &  & \\
        & \textbf{LPR} & $-0.72$*** &  $-0.23$* &  \\
        & \textbf{WPR} & $0.01$	& $-0.31$	 & $0.08$ \\
        \midrule
        \multirow{3}{*}{\textbf{Crosslingual}} &
       \textbf{ $H_{\mathbf{C}}$[W]} & $0.54$*** &  & \\
        & \textbf{LPR}& $\mathbf{-0.87}$*** &  $-0.38$*** &  \\
        & \textbf{WPR} & $-0.27$**	& $-0.47$***	 & $\mathbf{0.37}$*** \\
        \bottomrule
        
    \end{tabular}
    }
\caption{Spearman correlation between Language Confusion Entropy and Pass Rates at both Word level and Line level with LCB. The strongest correlation is in \textbf{bold}.}    
\label{tab:lcb_pass_entropy}
\end{table}

\begin{figure}[t!]
    \centering
    \includegraphics[width=\linewidth]{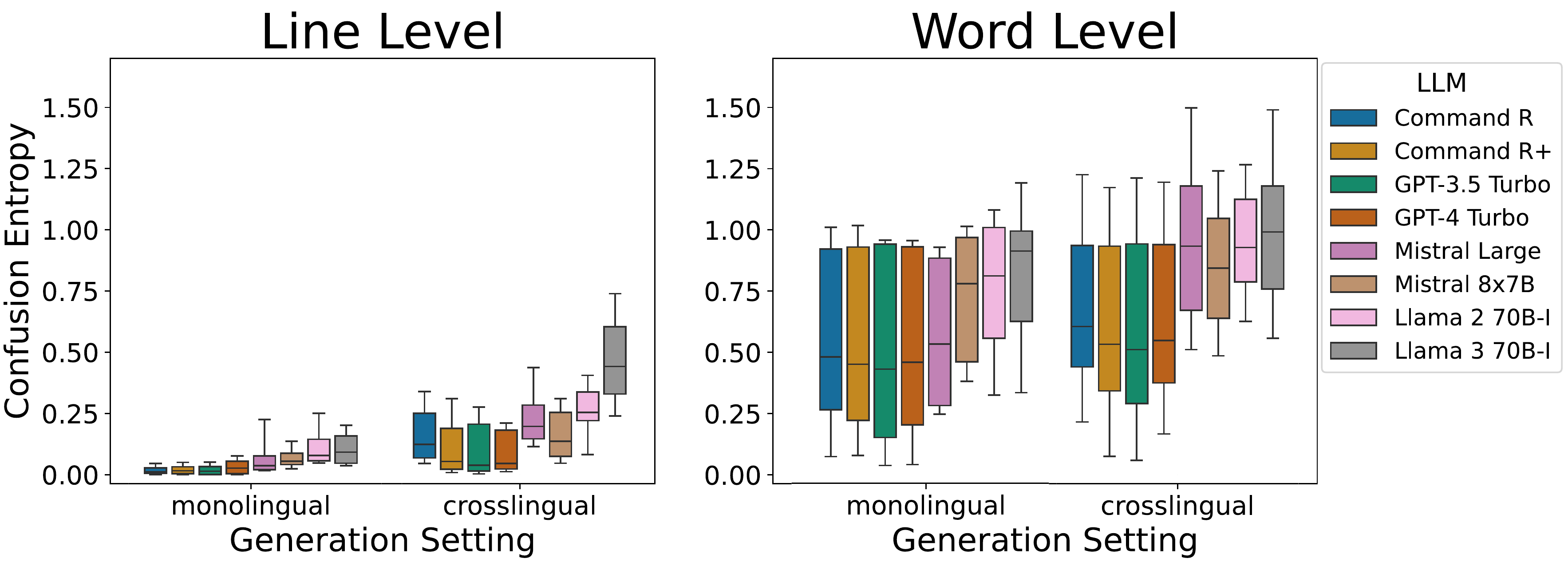}
    \caption{Language confusion entropy for LCB across generation settings by LLMs at line and word Level.}
    \label{fig:lcb_entropy_llms}
\end{figure}
\begin{figure*}[t!]
    \centering
        \includegraphics[width=0.8\linewidth]{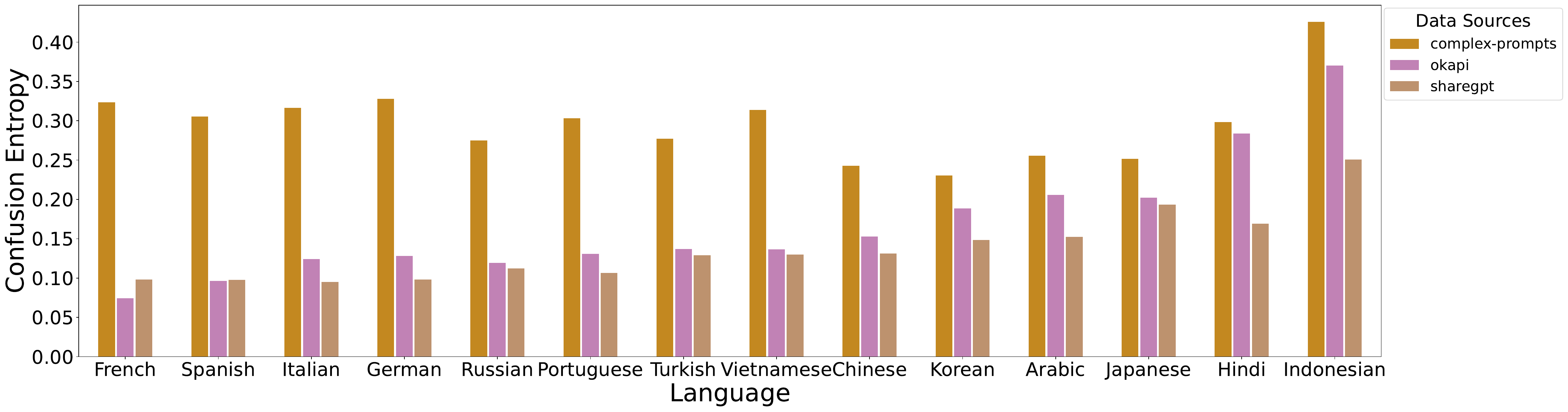}
\caption{Language confusion for \textbf{LCB}  across data sources at line level for crosslingual setting.}
\label{fig:lcb_data_source_langs}
\end{figure*}

\begin{figure}[t!]
    \centering
    \includegraphics[width=\linewidth]{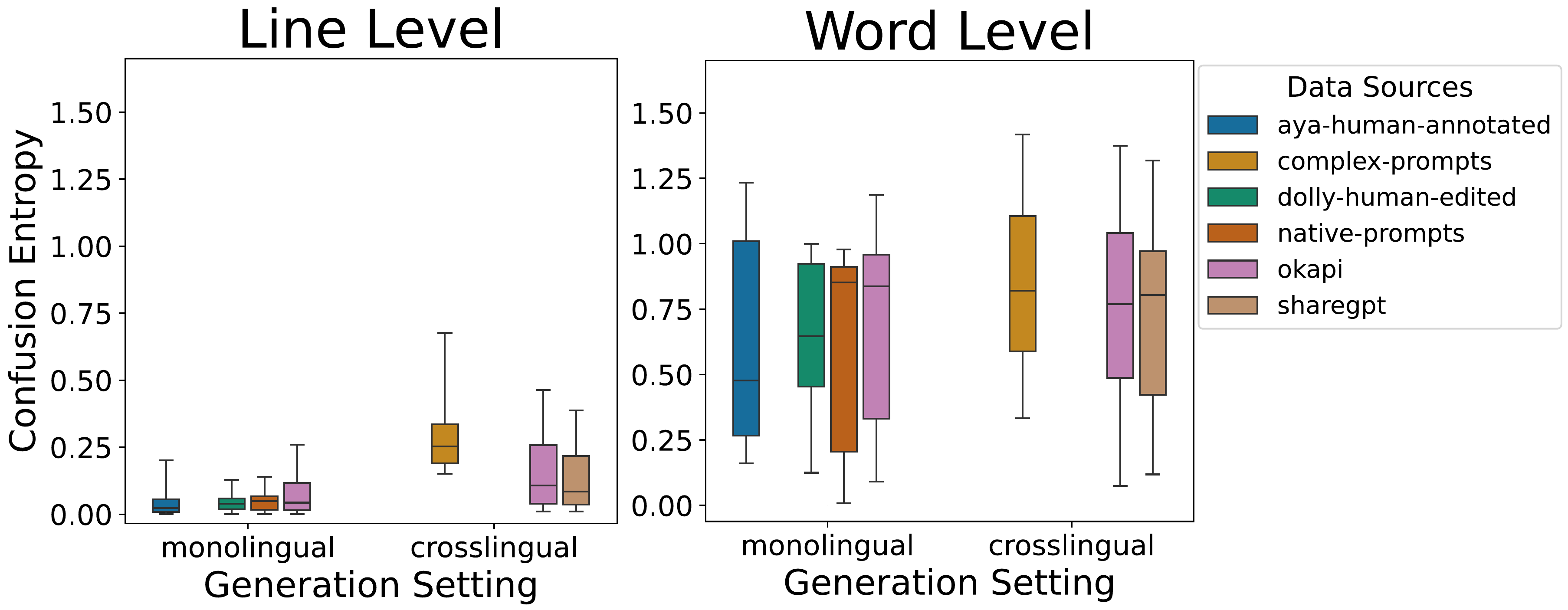}
       \caption{Language Confusion for LCB across generation settings by data sources at line and word Level.}

    \label{fig:lcb_entropy_source}
\end{figure}

\paragraph{Language Confusion Entropy Across LLMs}

As shown in Table~\ref{tab:lcb_confusion_entropy_llms} and Fig.~\ref{fig:lcb_entropy_llms}, language confusion is more likely to occur in crosslingual compared to monolingual,
with each LLM presenting significant variance.
Word-level language confusion presents more variance per LLM and higher severity than line-level.
Overall, it is consistent that Command and GPT LLMs have relatively lower language confusion than Mistral and LLaMA LLMs, projecting similar findings from~\citet{marchisio2024understanding}.

There is a clear consistency in language confusion across different LLMs, particularly at the line level and in crosslingual settings, as shown in Fig.~\ref{fig:lcb_entropy_llms}.
LLaMA 3 70B-I consistently exhibits the highest confusion across nearly all languages, while GPT-4 Turbo demonstrates the lowest confusion, especially for high-resource languages like French, Spanish, German, and Chinese. Command R+ also shows relatively low confusion across most languages, except Indonesian.

Notably, languages that are written in non-Latin scripts (on the right side of the X-axis), such as Vietnamese, Chinese, Korean, Arabic, Japanese, Hindi, and Indonesian, consistently show higher confusion entropies across most LLMs, especially in the LLaMA models. In contrast, Latin-script languages like French, Spanish, German, and Italian tend to have lower confusion rates across all LLMs, particularly in GPT-4 Turbo and Command R+.

\paragraph{Language Confusion Entropy Across Data Sources}
The data for monolingual and crosslingual tasks in \textbf{LCB} consists of 2,600 and 4,500 prompts, respectively, sourced from 7 datasets (details in Table~\ref{tab:lcb_datasources}).
As shown in Fig.~\ref{fig:lcb_data_source_langs}, language confusion is more pronounced in crosslingual settings and at the word level. Monolingually, language confusion tends to align with the median word length (W) of prompts in each dataset. For example, Aya, Dolly, Okapi, and Native prompts have median lengths of 9, 10, 13, and 19 words, respectively, and their language confusion follows this order.
Crosslingually, the Complex Prompts dataset has the highest median word length (159, compared to 18 for ShareGPT and 15 for Okapi), and it also exhibits the highest language confusion.

Observing language confusion across datasets at the line level for crosslingual settings (Fig.~\ref{fig:lcb_entropy_source}),\footnote{We use ``Language Confusion Entropy'' and ``Confusion Entropy'' interchangeably in this paper.} a clear pattern emerges. Complex Prompts has the highest confusion across all languages, while ShareGPT shows the lowest confusion for most languages, except for French and Spanish.
Consistent with previous findings, languages written in non-Latin scripts show higher confusion in datasets like Okapi and ShareGPT. However, in Complex Prompts, non-Latin-script languages such as Chinese, Korean, Arabic, and Japanese demonstrate lower confusion than Latin-script languages.

\begin{figure*}[t!]
    \centering
    \includegraphics[width=0.85\linewidth]{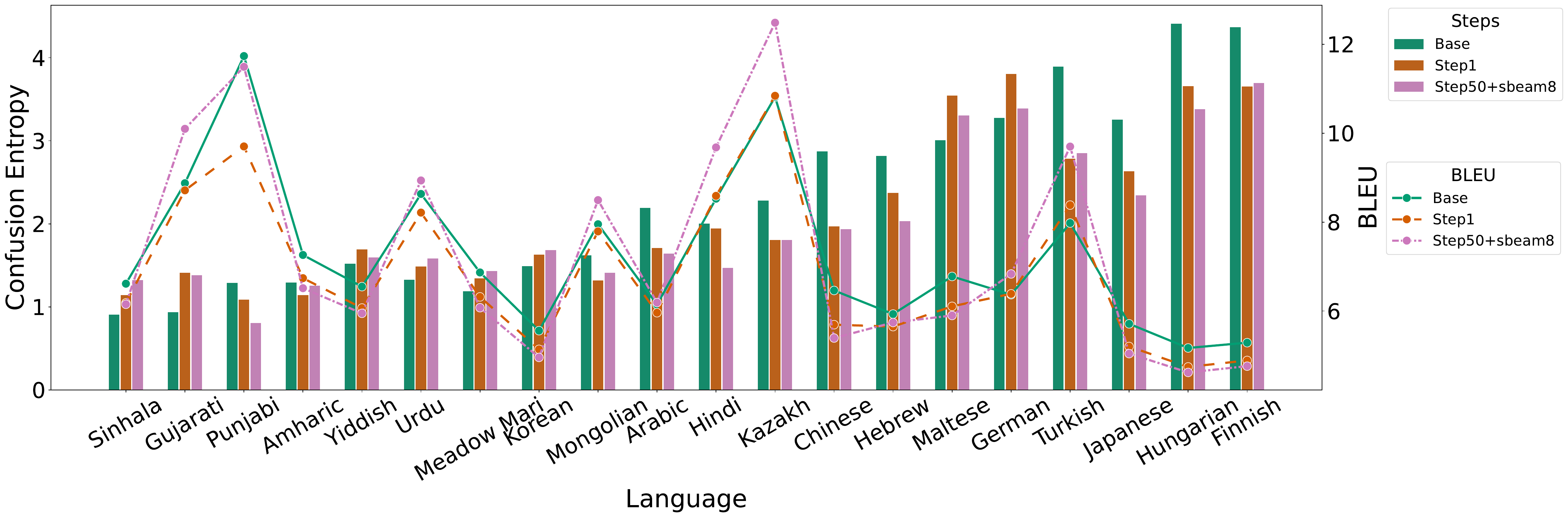}
     \includegraphics[width=0.85\linewidth]{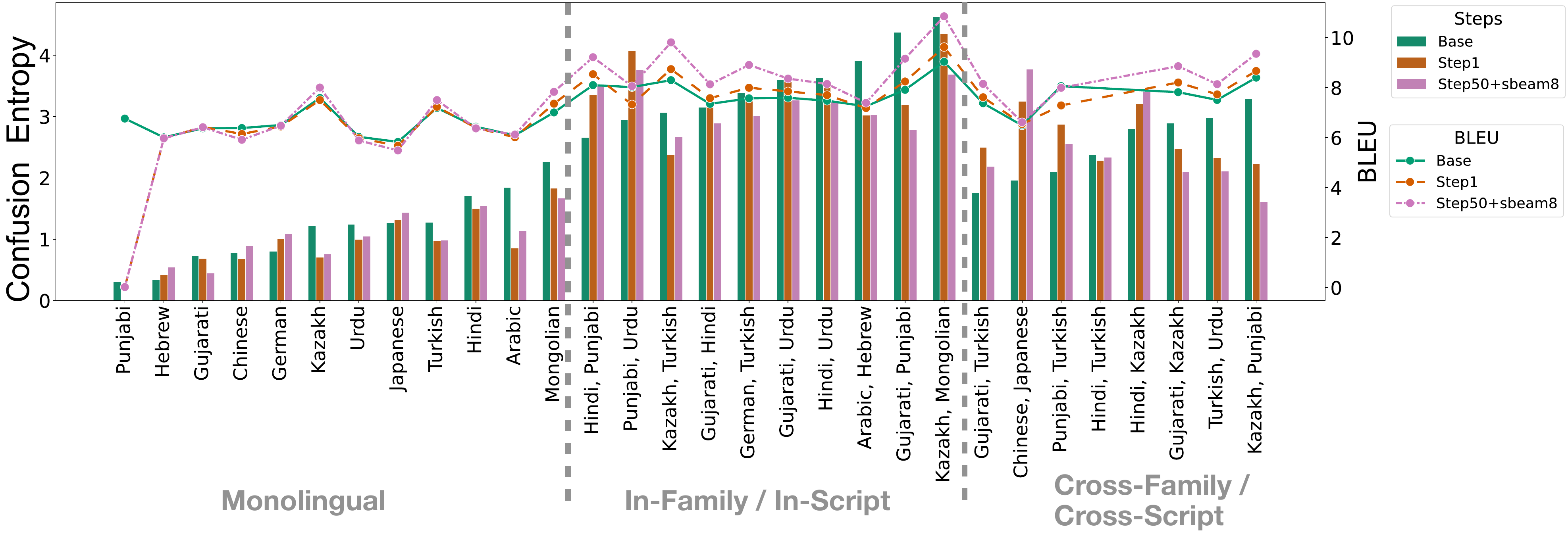}
    \caption{Language Confusion and Text Reconstruction Performance (BLEU) in Multilingual Textual Embedding Inversion Attacks at Line Level for Crosslingual Settings for Eval (Top) and Train (Bottom) Languages.}
    \label{fig:language_confusion_inversion}
\end{figure*}

\subsection{Language Confusion in Multilingual Textual Embedding Inversion Security}~\label{subsec:inversion_language_confusion}
\noindent\textbf{Language Confusion Entropy for Eval Languages}
When embeddings are in languages that are more likely to be confused, they are more prone to being inverted into text in "incorrect" languages, reducing the inversion performance, especially with word-matching metrics like BLEU~\citep{post2018call}. Also, the languages generated by the inversion model are often skewed by the pre-training data of the LLM, such as mT5 in \textbf{MTEI}.

\begin{table}[ht!]
    \centering
     \resizebox{0.85\linewidth}{!}{ 
    \begin{tabular}{c|c|cccc}
    \toprule 
        & & \textbf{$H_{\mathbf{C}}$[L]} &   \textbf{$H_{\mathbf{C}}$[W]} & \textbf{BLEU}   \\
        \midrule
                \multirow{3}{*}{\textbf{All}} &
        \textbf{$H_{\mathbf{C}}$[W]} & $0.89$*** &  & \\
       &  \textbf{BLEU} & $\mathbf{-0.62}$***	& $ \mathbf{-0.44}$***& \\
      &  \textbf{mT5} & $0.71$***	& $0.62$*** &	$-0.32$*	\\
   
        \midrule
        
        \multirow{3}{*}{\textbf{Monolingual}} &
         \textbf{$H_{\mathbf{C}}$[W]} & $0.75$*** &  & \\
       & \textbf{BLEU} & $0.25$&	$0.17$ & \\
      &  \textbf{mT5} & 0.1&	0.26&	$-0.59$***	\\

         \midrule
   \multirow{3}{*}{\textbf{Crosslingual}} &
        \textbf{$H_{\mathbf{C}}$[W]} & $\mathbf{0.9}$*** &  & \\
       & \textbf{BLEU} & $-0.48$***&	$-0.36$** & \\
      &  \textbf{mT5} & $\mathbf{0.76}$***&	$\mathbf{0.66}$***&	$-0.32$*	\\

        \bottomrule
    \end{tabular}
    }
 \caption{Spearman Correlations among $H_{\mathbf{C}}$ at Line Level and Word Level and BLEU score for \textbf{MTEI}, and the percentage of pre-training data in mT5 for eval languages. Strongest correlations are in \textbf{bold}.}  
\label{tab:inversion_entropy_bleu_eval}
\end{table}


To test this intuition, we calculate the Spearman correlation among language confusion entropy, BLEU scores, and the percentage of respective languages in the pre-training data of mT5 (see Table~\ref{tab:inversion-languages} for details).
As shown in Table~\ref{tab:inversion_entropy_bleu_eval}, for eval languages, 
\textbf{$H_{\mathbf{C}}$[L]} is strongly correlated with inversion performance across generation settings, confirming that language confusion negatively impacts reconstruction performance.

Moreover, \textbf{$H_{\mathbf{C}}$[L]} and \textbf{$H_{\mathbf{C}}$[W]} are both strongly correlated with the proportion of languages in pre-training data of mT5, particularly in crosslingual setting.
This indicates that languages with higher representation in the pre-training data are more prone to confusion, both at the word and line levels.
This observation is further validated by empirical evidence: when an inversion model trained on Arabic is used to invert embeddings in Meadow Mari (unseen in mT5), the model often generates English, highlighting the influence of pre-training data on embedding inversion attacks. Fig~\ref{fig:language_confusion_inversion} (top) provides a visualization of language confusion at the line level for each language, alongside their corresponding reconstruction performance in BLEU at each step of the evaluation in the crosslingual inversion attacks. It shows directly that lower-resourced languages present lower confusion, especially for non-Latin script languages, whereas they are also more vulnerable in terms of higher reconstruction performance, for example, Gujarati, Punjabi, and Urdu.

\begin{table*}[t!]
    \centering
      \resizebox{\linewidth}{!}{ 
    \begin{tabular}{l|>{\columncolor{gray!20}}c|cc|cc|cc||>{\columncolor{gray!20}}c|cc|cc|cc}
    \toprule
    
    \multirow{3}{*}{\textbf{Language Graph}}& \multicolumn{7}{|c||}{\textbf{LCB}}   & \multicolumn{7}{c}{\textbf{Textual Embedding Inversion}} \\
        &  \textbf{AVG} & \multicolumn{2}{|c|}{\textbf{ALL}} &\multicolumn{2}{|c|}{\textbf{Monolingual}} &\multicolumn{2}{|c||}{\textbf{Crosslingual}} & \textbf{AVG} &  \multicolumn{2}{c|}{\textbf{ALL}} &\multicolumn{2}{|c|}{\textbf{Monolingual}} &\multicolumn{2}{|c}{\textbf{Crosslingual}} \\
        
         & & Line & Word &  Line & Word  &  Line & Word  & & Line & Word  &  Line & Word  &  Line & Word \\ 
         \midrule
           
        \textbf{Grambank} &0.2650&  0.1830 & 0.2827 & 0.2726 & 0.4042 & 0.1791 & 0.2682 & 0.7538 & \textbf{0.8005} & 0.7891 & 0.7504 & 0.5434 & \textbf{0.8230} & 0.8162 \\ 
        \textbf{WALS} &0.1947 & 0.0420&	0.2908	& 0.0816 &0.4286&  0.0388 &	0.2865 & 0.7377 & 0.8722 & \textbf{0.7854} & 0.6618 & 0.4102 & \underline{0.8803} & 0.8164 \\ 
        
        \textbf{WALS $\backslash$ Phon.} & 0.1892&  0.0409 &	0.2889&	0.0799&	0.4253	& 0.0379 &	0.2620 & \underline{0.7360} & 0.8768 & \underline{0.7880} & 0.6495 & 0.4036 & 0.8837 & \underline{0.8147} \\

        \midrule
    \underline{\textbf{Lang2Vec}} &  &   &    &   &  &   & &   & &  &   &  &  &   \\       
\textbf{Inventory}  & 0.1650 &0.0812 & \textbf{0.2003} & 0.1374 & \textbf{0.3114} & 0.0637 & \textbf{0.1961} &0.8154 & 0.9678 & 0.8410 & 0.8225 & 0.4124 & 0.9720 & 0.8768 \\ 
        \textbf{Syntactic} &0.2925&  0.1949 & 0.3771 & 0.2244 & 0.4679 & 0.1505 & 0.3405 & 0.8651& 1.0179 & 0.8668 & 0.8872 & 0.5447 & 1.0001 & 0.8742 \\ 
        \textbf{Phonological} &0.2260& 0.1009 & 0.3126 & 0.1552 & 0.4235 & 0.0830 & 0.2808 & 1.3161& 1.7178 & 1.6133 & 0.6859 & 0.6004 & 1.6599 & 1.6191 \\ 
        \textbf{Genetic} & 12.8307 & 13.4548 & 12.2278 & 12.9350 & 12.3563 & 13.5850 & 12.4249 & 14.9443& 14.6988 & 14.9756 & 18.5227 & 13.5734 & 13.8721 & 14.0232 \\ 
        \midrule
        \textbf{ASJP SVD} & 0.5859 & \underline{0.0253} & 1.0597 & \textbf{0.0522} & 1.0395 & \underline{0.0225} & 1.3164 & 2.8254 & 2.9730 & 3.3461 & 0.0820 & 1.6568 & 3.8882 & 5.0063 \\ 
        \textbf{ASJP UMAP} & 0.9318 & 0.0994 & 1.6767 & 0.1290 & 1.6055 & 0.0946 & 1.9856 & 3.3217 & 3.5364 & 4.0245 & 0.0811 & 2.3479 & 4.3690 & 5.5711 \\ 
             \midrule

     \underline{\textbf{Colex2Lang}} &  &   &    &   &  &   & &   & &  &   &  &  &   \\
          
        \textbf{CLICS} & 0.1665 & 0.0289 & 0.2522 & \underline{0.0589} & 0.3776 & 0.0260 & 0.2490 & \textbf{0.7333} & 0.9492 & 0.8335 & 0.3690 & \textbf{0.2672} & 0.9503 & 0.8693 \\ 
        \textbf{WN} & \textbf{0.1489} & \textbf{0.0242} & \underline{0.2154} & \textbf{0.0522} & \underline{0.3404} & \textbf{0.0218} & \underline{0.2149} & 0.7794 & 1.0105 & 0.8892 & 0.3263  & \underline{0.3894} & 1.0445 & 0.9454 \\ 
        \textbf{WN\_CONCEPT} & \underline{0.1490} & \textbf{0.0242} & 0.2175 & 0.0522 & 0.3426 & \textbf{0.0218} & 0.2168 & 0.7791 & 1.0100 & 0.8836 & \textbf{0.3263} & \underline{0.3894} & 1.0427 & 0.9390 \\ 
       
    

        \bottomrule 
    \end{tabular}
    }
    \caption{KL Divergence between Language Similarity Graphs and the Language Confusion Matrices for Target/ Eval Languages from LCB and Inversion Tasks. The best results (lowest) are \textbf{bolded}, and the second best are \underline{underlined}.
    \label{tab:langsim}}
\end{table*}

\paragraph{Language Confusion Entropy for Train Languages}
In~\textbf{MTEI}, the inversion models are trained in three different settings - monolingual, in-family and in-script, and cross-family and cross-script.
As shown in Fig.~\ref{fig:language_confusion_inversion} (bottom) and Table~\ref{tab:mtei_lc_bleu_train}, monolingual training renders lower language confusion for each train language while pairing training languages, in-script/in-family training renders higher language confusion compared to cross-script/cross-family training. 
These findings substantiate the intuition that similar languages are more prone to confusion.

Our study reveals that inversion performance significantly improves when trained in in-script/in-family settings (ref. Table~\ref{tab:mtei_lc_bleu_train} in the Appendix). Crosslingual inversion performances are comparable to in-script/in-family training when trained in Kazakh (Latin-script) combined with Gujarati and Punjabi, respectively, and language confusion is notably lower. This suggests that while similar languages tend to increase confusion, certain crosslingual combinations can achieve strong performance without the added confusion seen in in-family/in-script training.
Overall, these findings highlight the trade-off between inversion performance and language confusion, indicating further optimization is needed to strike the ideal balance between them.

\subsection{Language Confusion and  Linguistic Typology}

Table~\ref{tab:langsim} shows the best results from KL divergence between language confusion and language similarities based on different language graphs for both \textbf{LCB} and \textbf{MTEI}, using Algorithm~\ref{alg:kl_divergence_matrices}, the whole results are presented in Table~\ref{tab:kl_all} in Appendix.

Our findings reveal strong correlations between language confusion and language similarities based on various typological sources.
For instance, the similarity measures based on semantic typology correlate the most strongly, followed closely by more general lexical similarity measures.
Language similarities based on typological feature databases like Grambank and WALS show stronger correlations than those based on parallel Bible texts~\citep{ostling-tiedemann-2017-continuous}. Interestingly, and echoing previous findings on typological variation, we find genetic variation is a poor proxy for this analysis~\citep{bjerva-etal-2019-language,ploeger2024typological}, indicating the need for theoretically grounded approaches to linguistic interpretation.



\section{Discussion}

\paragraph{Language Confusion Entropy~\textbf{$H_{\mathbf{C}}$} vs. Pass Rate (PR)} 
As shown in Section~\ref{sec:lc_llm_prompting}, $H_{\mathbf{C}}$ and PR are correlated but measure distinct aspects of language generation. Importantly,~\textbf{$H_{\mathbf{C}}$} is not intended to replace PR but to offer a deeper quantitative characterization of language confusion.

\textbf{$H_{\mathbf{C}}$} quantifies behavioral uncertainty through the entropy of a model's probability distribution across languages. It captures partial confusions (e.g., 40\% confidence in the target language, 60\% spread across others), even when outputs technically ``pass.'' \textbf{PR} is a binary performance metric assessing whether outputs meet target language requirements.

Hence, $H_{\mathbf{C}}$ complements PR by revealing hidden confusion patterns: a model with high PR could still exhibit high $H_{\mathbf{C}}$ if probabilities scatter across non-target languages. This makes $H_{\mathbf{C}}$ particularly valuable for diagnosing \textit{how} models arrive at correct outputs, not just whether they do.

\paragraph{Research Questions}
In response to \textbf{RQ1}, we proposed an effective metric \textit{Language Confusion Entropy}, through which we identified several patterns contributing to language confusion. These include prompt complexity, imbalanced distributions of training sources, and language similarities - all play a significant role in language confusion. Furthermore, our findings indicate that these factors strongly correlate with inversion performance and the pretraining languages in LLMs.

Inherent vulnerabilities in LLMs stem from intrinsic design, training processes, or model architectures, which are not directly caused by attacks or improper use. 
However, external attacks can amplify and leverage the vulnerabilities, which are directly reflected in embedding inversion attacks. Further, decoding unexpected languages disrupts the user experience. We have discussed that i) language confusion stems from training data inequality, ii) the phenomena is pervasive from related works, iii) language confusion is well-captured by the proposed LCE and shows consistent trends across different architectures of LLMs and datasets and experimental settings, and (iv) can be influenced and explained by linguistic typological variations. 
These affords a positive response to~\textbf{RQ2}.

Moreover, there is a potential to leverage language similarities as a prior for LLM alignment and security.
For instance, Table~\ref{tab:langsim} shows that the language similarities from colexification patterns afford a strong correlation with language confusion (with low KL divergence), which indicates that LLMs easily confuse languages that contain words that are crosslingually capturing the same senses. When an LLM is exposed to multilingual data with more distinct colexification patterns, it could enhance its ability to distinguish them and make them more resilient against language confusion.
This strategy could promote more resilient LLMs, as we have shown that models are less likely to confuse typologically dissimilar languages. Hence, exploring typology-aware design strategies could provide both offensive and defensive insights in LLM security.

\paragraph{Potential Misuse}
 The Language confusion metric identifies uncertainty in language identification, highlighting areas where LLMs may be prone to errors. Our findings show that language similarities correlate with language confusion patterns in LLMs. When similar languages lack robust safety measures compared to well-protected high-resource languages, they could be exploited for crosslingual attacks in a more targeted manner, such as backdoors and jailbreaking \citep{gcg,li2024faster}. Additionally, the ability of LLMs to switch between languages may pose risks where safety measures aren't consistently implemented across all languages. 
 The Language confusion metric identifies uncertainty in language identification, highlighting areas where LLMs may be prone to errors. Our findings show that language similarities correlate with language confusion patterns in LLMs. When similar languages lack robust safety measures compared to well-protected high-resource languages, they could be exploited for crosslingual attacks in a more targeted manner, such as jailbreaking \citep{he_tuba_2024} and backdoors \citep{wang_backdoor_2024}. Additionally, the ability of LLMs to switch between languages may pose risks where safety measures aren't consistently implemented across all languages. 
 It has been demonstrated that LLMs are particularly vulnerable to crosslingual attacks in related work~\citep{wang_backdoor_2024,he_tuba_2024} and more recently~\citep{poppi2024understandingfragilitymultilingualllms}, mainly because the conventional defense mechanisms designed for monolingual settings are ineffective in multilingual settings. To enhance LLM security, adversarial exploits can be detected by monitoring high language confusion entropy, especially in crosslingual settings. 
 Our work suggests that language confusion quantification and its connection to language similarities can be leveraged to raise awareness of such vulnerabilities, and also provide potential revenue for developing mitigation strategies.
 

\section{Conclusion and Future Work}

Addressing the challenge of language confusion, we introduce \textit{Language Confusion Entropy}, a novel metric that quantifies language confusion by re-weighting language distributions and emphasizing long-tail patterns. 
This metric captures language confusion in multilingual LLM tasks, revealing patterns of uncertainty in both training and evaluation phases. 
Our findings show strong correlations between language confusion and semantic similarities among languages, with less confusion observed in low-resource languages and when training incorporates diverse scripts and language families. These insights confirm that language confusion fundamentally impacts LLMs and suggest linguistic typology as a potential tool for enhancing model security. 
Detecting language confusion enables smoother and more precise interactions in multilingual contexts while enhancing user trust in LLM-based AI systems across domains such as legal services and healthcare. 
In future work, we aim to apply these findings to practical applications, such as developing typology-aware defense to improve LLM alignment and security.
Key applications include cross-lingual chatbots, translation services, detecting code-switching, and improving multilingual speech recognition by reducing ambiguity.





\section*{Limitations}
A core limitation of this work is that some of our analysis and downstream implications can only be carried out on languages that are represented in typological databases.
When part of the work, limited to typological databases, inspires downstream solutions in terms of defense mechanisms, undocumented languages may not benefit from these advances. 
However, we have also increased coverage of languages in constructing language graphs using data-driven methods. 
Our core method is also not limited to any typological database.

\section*{Ethics Statement}
This work adheres to the ACL ethics guidelines. 
We investigate language confusion and link findings to security vulnerabilities of low-resource languages, including those using non-Latin scripts and with diverse typologies.
The potential misuse has been extensively discussed.
Our work highlights how these factors can be used to potentially improve the security of low-resource language technology.
We encourage the community to incorporate a broader range of languages in NLP security research, to ensure that low-resource languages are also covered by defense mechanisms developed in the future.

\section*{Acknowledgements}
YC and JB are funded by the Carlsberg Foundation, under the \textit{Semper Ardens: Accelerate} programme (project nr. CF21-0454). 

\bibliography{main}

\begin{thebibliography}{66}
\providecommand{\natexlab}[1]{#1}

\bibitem[{Achiam et~al.(2023)Achiam, Adler, Agarwal, Ahmad, Akkaya, Aleman, Almeida, Altenschmidt, Altman, Anadkat et~al.}]{achiam2023gpt}
Josh Achiam, Steven Adler, Sandhini Agarwal, Lama Ahmad, Ilge Akkaya, Florencia~Leoni Aleman, Diogo Almeida, Janko Altenschmidt, Sam Altman, Shyamal Anadkat, et~al. 2023.
\newblock Gpt-4 technical report.
\newblock \emph{arXiv preprint arXiv:2303.08774}.

\bibitem[{Augenstein et~al.(2024)Augenstein, Baldwin, Cha, Chakraborty, Ciampaglia, Corney, DiResta, Ferrara, Hale, Halevy et~al.}]{augenstein2024factuality}
Isabelle Augenstein, Timothy Baldwin, Meeyoung Cha, Tanmoy Chakraborty, Giovanni~Luca Ciampaglia, David Corney, Renee DiResta, Emilio Ferrara, Scott Hale, Alon Halevy, et~al. 2024.
\newblock Factuality challenges in the era of large language models and opportunities for fact-checking.
\newblock \emph{Nature Machine Intelligence}, pages 1--12.

\bibitem[{Bird and Loper(2004)}]{bird-loper-2004-nltk}
Steven Bird and Edward Loper. 2004.
\newblock \href {https://aclanthology.org/P04-3031} {{NLTK}: The natural language toolkit}.
\newblock In \emph{Proceedings of the {ACL} Interactive Poster and Demonstration Sessions}, pages 214--217, Barcelona, Spain. Association for Computational Linguistics.

\bibitem[{Bjerva(2024)}]{bjerva-2024-role}
Johannes Bjerva. 2024.
\newblock \href {https://doi.org/10.1162/coli_a_00498} {The role of typological feature prediction in {NLP} and linguistics}.
\newblock \emph{Computational Linguistics}, 50(2):781--794.

\bibitem[{Bjerva et~al.(2019{\natexlab{a}})Bjerva, Kementchedjhieva, Cotterell, and Augenstein}]{bjerva-etal-2019-probabilistic}
Johannes Bjerva, Yova Kementchedjhieva, Ryan Cotterell, and Isabelle Augenstein. 2019{\natexlab{a}}.
\newblock \href {https://doi.org/10.18653/v1/N19-1156} {A probabilistic generative model of linguistic typology}.
\newblock In \emph{Proceedings of the 2019 Conference of the North {A}merican Chapter of the Association for Computational Linguistics: Human Language Technologies, Volume 1 (Long and Short Papers)}, pages 1529--1540, Minneapolis, Minnesota. Association for Computational Linguistics.

\bibitem[{Bjerva et~al.(2019{\natexlab{b}})Bjerva, {\"O}stling, Veiga, Tiedemann, and Augenstein}]{bjerva-etal-2019-language}
Johannes Bjerva, Robert {\"O}stling, Maria~Han Veiga, J{\"o}rg Tiedemann, and Isabelle Augenstein. 2019{\natexlab{b}}.
\newblock \href {https://doi.org/10.1162/coli_a_00351} {What do language representations really represent?}
\newblock \emph{Computational Linguistics}, 45(2):381--389.

\bibitem[{Brown(2020)}]{brown2020language}
Tom~B Brown. 2020.
\newblock Language models are few-shot learners.
\newblock \emph{arXiv preprint arXiv:2005.14165}.

\bibitem[{Chen et~al.(2023{\natexlab{a}})Chen, Ma, Zhang, Wei, and Chang}]{chen-etal-2023-target}
Liang Chen, Shuming Ma, Dongdong Zhang, Furu Wei, and Baobao Chang. 2023{\natexlab{a}}.
\newblock \href {https://doi.org/10.18653/v1/2023.findings-acl.608} {On the off-target problem of zero-shot multilingual neural machine translation}.
\newblock In \emph{Findings of the Association for Computational Linguistics: ACL 2023}, pages 9542--9558, Toronto, Canada. Association for Computational Linguistics.

\bibitem[{Chen et~al.(2022)Chen, Song, Wu, Wang, Xu, Chen, Zhou, and Li}]{chen2022mtg}
Yiran Chen, Zhenqiao Song, Xianze Wu, Danqing Wang, Jingjing Xu, Jiaze Chen, Hao Zhou, and Lei Li. 2022.
\newblock Mtg: A benchmark suite for multilingual text generation.
\newblock In \emph{Findings of the Association for Computational Linguistics: NAACL 2022}, pages 2508--2527.

\bibitem[{Chen et~al.(2023{\natexlab{b}})Chen, Biswas, and Bjerva}]{chen-etal-2023-colex2lang}
Yiyi Chen, Russa Biswas, and Johannes Bjerva. 2023{\natexlab{b}}.
\newblock \href {https://aclanthology.org/2023.nodalida-1.67} {{C}olex2{L}ang: Language embeddings from semantic typology}.
\newblock In \emph{Proceedings of the 24th Nordic Conference on Computational Linguistics (NoDaLiDa)}, pages 673--684, T{\'o}rshavn, Faroe Islands. University of Tartu Library.

\bibitem[{Chen et~al.(2025)Chen, Biswas, Lent, and Bjerva}]{chen2024oddsovercomingtypologyscript}
Yiyi Chen, Russa Biswas, Heather Lent, and Johannes Bjerva. 2025.
\newblock \href {https://arxiv.org/abs/2408.11749} {Against all odds: Overcoming typology, script, and language confusion in multilingual embedding inversion attacks}.
\newblock In \emph{The 39th Annual AAAI Conference on Artificial Intelligence}.

\bibitem[{Chen and Bjerva(2023)}]{chen-bjerva-2023-colexifications}
Yiyi Chen and Johannes Bjerva. 2023.
\newblock \href {https://doi.org/10.18653/v1/2023.sigmorphon-1.11} {Colexifications for bootstrapping cross-lingual datasets: The case of phonology, concreteness, and affectiveness}.
\newblock In \emph{Proceedings of the 20th SIGMORPHON workshop on Computational Research in Phonetics, Phonology, and Morphology}, pages 98--109, Toronto, Canada. Association for Computational Linguistics.

\bibitem[{Chen et~al.(2024)Chen, Lent, and Bjerva}]{chen-etal-2024-text}
Yiyi Chen, Heather Lent, and Johannes Bjerva. 2024.
\newblock \href {https://doi.org/10.18653/v1/2024.acl-long.422} {Text embedding inversion security for multilingual language models}.
\newblock In \emph{Proceedings of the 62nd Annual Meeting of the Association for Computational Linguistics (Volume 1: Long Papers)}, pages 7808--7827, Bangkok, Thailand. Association for Computational Linguistics.

\bibitem[{Chirkova and Nikoulina(2024)}]{chirkova-nikoulina-2024-key}
Nadezhda Chirkova and Vassilina Nikoulina. 2024.
\newblock \href {https://doi.org/10.18653/v1/2024.naacl-long.401} {Key ingredients for effective zero-shot cross-lingual knowledge transfer in generative tasks}.
\newblock In \emph{Proceedings of the 2024 Conference of the North American Chapter of the Association for Computational Linguistics: Human Language Technologies (Volume 1: Long Papers)}, pages 7222--7238, Mexico City, Mexico. Association for Computational Linguistics.

\bibitem[{Collin(2010)}]{collin2010ethnologue}
Richard~Oliver Collin. 2010.
\newblock Ethnologue.
\newblock \emph{Ethnopolitics}, 9(3-4):425--432.

\bibitem[{Collins and Kayne(2011)}]{collins2011syntactic}
Chris Collins and Richard Kayne. 2011.
\newblock Syntactic structures of the world’s languages (sswl).

\bibitem[{Deng et~al.(2024)Deng, Zhang, Pan, and Bing}]{deng_multilingual_2024}
Yue Deng, Wenxuan Zhang, Sinno~Jialin Pan, and Lidong Bing. 2024.
\newblock \href {https://arxiv.org/abs/2310.06474 [cs]} {Multilingual jailbreak challenges in large language models}.
\newblock \emph{Preprint}, arxiv:2310.06474 [cs].

\bibitem[{Di~Natale et~al.(2021)Di~Natale, Pellert, and Garcia}]{di2021colexification}
Anna Di~Natale, Max Pellert, and David Garcia. 2021.
\newblock Colexification networks encode affective meaning.
\newblock \emph{Affective Science}, 2(2):99--111.

\bibitem[{Fellbaum(2010)}]{fellbaum2010wordnet}
Christiane Fellbaum. 2010.
\newblock Wordnet.
\newblock In \emph{Theory and applications of ontology: computer applications}, pages 231--243. Springer.

\bibitem[{Fran{\c{c}}ois(2008)}]{franccois2008semantic}
Alexandre Fran{\c{c}}ois. 2008.
\newblock Semantic maps and the typology of colexification.
\newblock \emph{From polysemy to semantic change: Towards a typology of lexical semantic associations}, 106:163.

\bibitem[{Guo et~al.(2024)Guo, Ren, Hu, Cao, Li, and Huang}]{guo2024steering}
Ping Guo, Yubing Ren, Yue Hu, Yanan Cao, Yunpeng Li, and Heyan Huang. 2024.
\newblock Steering large language models for cross-lingual information retrieval.
\newblock In \emph{Proceedings of the 47th International ACM SIGIR Conference on Research and Development in Information Retrieval}, pages 585--596.

\bibitem[{Haspelmath(2008)}]{haspelmath2008typological}
Martin Haspelmath. 2008.
\newblock The typological database of the world atlas of language structures.
\newblock \emph{Typological databases. Mouton de Gruyter, Berlin}.

\bibitem[{He et~al.(2024)He, Wang, Xu, Minervini, Stenetorp, Rubinstein, and Cohn}]{he_tuba_2024}
Xuanli He, Jun Wang, Qiongkai Xu, Pasquale Minervini, Pontus Stenetorp, Benjamin I.~P. Rubinstein, and Trevor Cohn. 2024.
\newblock \href {https://arxiv.org/abs/2404.19597 [cs]} {{TuBA}: Cross-lingual transferability of backdoor attacks in {LLMs} with instruction tuning}.
\newblock \emph{Preprint}, arxiv:2404.19597 [cs].

\bibitem[{Iyer et~al.(2024)Iyer, Malik, Zhu, Stepachev, Chen, Haddow, and Birch}]{iyer-etal-2024-exploring}
Vivek Iyer, Bhavitvya Malik, Wenhao Zhu, Pavel Stepachev, Pinzhen Chen, Barry Haddow, and Alexandra Birch. 2024.
\newblock \href {https://doi.org/10.18653/v1/2024.americasnlp-1.25} {Exploring very low-resource translation with {LLM}s: The {U}niversity of {E}dinburgh{'}s submission to {A}mericas{NLP} 2024 translation task}.
\newblock In \emph{Proceedings of the 4th Workshop on Natural Language Processing for Indigenous Languages of the Americas (AmericasNLP 2024)}, pages 209--220, Mexico City, Mexico. Association for Computational Linguistics.

\bibitem[{Jiang et~al.(2024)Jiang, Sablayrolles, Roux, Mensch, Savary, Bamford, Chaplot, Casas, Hanna, Bressand et~al.}]{jiang2024mixtral}
Albert~Q Jiang, Alexandre Sablayrolles, Antoine Roux, Arthur Mensch, Blanche Savary, Chris Bamford, Devendra~Singh Chaplot, Diego de~las Casas, Emma~Bou Hanna, Florian Bressand, et~al. 2024.
\newblock Mixtral of experts.
\newblock \emph{arXiv preprint arXiv:2401.04088}.

\bibitem[{Joulin et~al.(2016)Joulin, Grave, Bojanowski, and Mikolov}]{joulin2016bag}
Armand Joulin, Edouard Grave, Piotr Bojanowski, and Tomas Mikolov. 2016.
\newblock Bag of tricks for efficient text classification.
\newblock \emph{arXiv preprint arXiv:1607.01759}.

\bibitem[{Karjus et~al.(2021)Karjus, Blythe, Kirby, Wang, and Smith}]{karjus2021conceptual}
Andres Karjus, Richard~A Blythe, Simon Kirby, Tianyu Wang, and Kenny Smith. 2021.
\newblock Conceptual similarity and communicative need shape colexification: An experimental study.
\newblock \emph{Cognitive Science}, 45(9):e13035.

\bibitem[{Kashyap(2019)}]{kashyap2019language}
Abhishek~Kumar Kashyap. 2019.
\newblock Language typology.
\newblock \emph{The Cambridge handbook of systemic functional linguistics}, pages 767--792.

\bibitem[{Kim et~al.(2022)Kim, Lee, and Oh}]{kim-etal-2022-toward}
Donggyu Kim, Garam Lee, and Sungwoo Oh. 2022.
\newblock \href {https://doi.org/10.18653/v1/2022.finnlp-1.4} {Toward privacy-preserving text embedding similarity with homomorphic encryption}.
\newblock In \emph{Proceedings of the Fourth Workshop on Financial Technology and Natural Language Processing (FinNLP)}, pages 25--36, Abu Dhabi, United Arab Emirates (Hybrid). Association for Computational Linguistics.

\bibitem[{Kullback and Leibler(1951)}]{kullback1951information}
Solomon Kullback and Richard~A Leibler. 1951.
\newblock On information and sufficiency.
\newblock \emph{The annals of mathematical statistics}, 22(1):79--86.

\bibitem[{Lai et~al.(2023)Lai, Nguyen, Ngo, Nguyen, Dernoncourt, Rossi, and Nguyen}]{lai-etal-2023-okapi}
Viet Lai, Chien Nguyen, Nghia Ngo, Thuat Nguyen, Franck Dernoncourt, Ryan Rossi, and Thien Nguyen. 2023.
\newblock \href {https://doi.org/10.18653/v1/2023.emnlp-demo.28} {Okapi: Instruction-tuned large language models in multiple languages with reinforcement learning from human feedback}.
\newblock In \emph{Proceedings of the 2023 Conference on Empirical Methods in Natural Language Processing: System Demonstrations}, pages 318--327, Singapore. Association for Computational Linguistics.

\bibitem[{Lee(2024)}]{kiwi-korean}
Minchul Lee. 2024.
\newblock \href {https://doi.org/10.23287/KJDH.2024.1.1.6} {Kiwi: Developing a korean morphological analyzer based on statistical language models and skip-bigram}.
\newblock \emph{Korean Journal of Digital Humanities}, 1(1).

\bibitem[{Levin and Oriyan(2018)}]{hebrew_tokenizer}
Yonti Levin and Daniel Oriyan. 2018.
\newblock Hebrew tokenizer.
\newblock \url{https://github.com/YontiLevin/Hebrew-Tokenizer}.
\newblock Accessed: 2024-10-15.

\bibitem[{Li and Murray(2023)}]{li-murray-2023-zero}
Tianjian Li and Kenton Murray. 2023.
\newblock \href {https://doi.org/10.18653/v1/2023.findings-acl.789} {Why does zero-shot cross-lingual generation fail? an explanation and a solution}.
\newblock In \emph{Findings of the Association for Computational Linguistics: ACL 2023}, pages 12461--12476, Toronto, Canada. Association for Computational Linguistics.

\bibitem[{Li et~al.(2024)Li, Li, Li, Lee, Cui, and Hu}]{li2024faster}
Xiao Li, Zhuhong Li, Qiongxiu Li, Bingze Lee, Jinghao Cui, and Xiaolin Hu. 2024.
\newblock Faster-gcg: Efficient discrete optimization jailbreak attacks against aligned large language models.
\newblock \emph{arXiv preprint arXiv:2410.15362}.

\bibitem[{Littell et~al.(2017)Littell, Mortensen, Lin, Kairis, Turner, and Levin}]{littell2017uriel}
Patrick Littell, David~R Mortensen, Ke~Lin, Katherine Kairis, Carlisle Turner, and Lori Levin. 2017.
\newblock Uriel and lang2vec: Representing languages as typological, geographical, and phylogenetic vectors.
\newblock In \emph{Proceedings of the 15th Conference of the European Chapter of the Association for Computational Linguistics: Volume 2, Short Papers}, pages 8--14.

\bibitem[{Lyu et~al.(2020)Lyu, He, and Li}]{Lyu2020DifferentiallyPR}
L.~Lyu, Xuanli He, and Yitong Li. 2020.
\newblock \href {https://api.semanticscholar.org/CorpusID:222134003} {Differentially private representation for nlp: Formal guarantee and an empirical study on privacy and fairness}.
\newblock \emph{ArXiv}, abs/2010.01285.

\bibitem[{Malaviya et~al.(2017)Malaviya, Neubig, and Littell}]{malaviya-etal-2017-learning}
Chaitanya Malaviya, Graham Neubig, and Patrick Littell. 2017.
\newblock \href {https://doi.org/10.18653/v1/D17-1268} {Learning language representations for typology prediction}.
\newblock In \emph{Proceedings of the 2017 Conference on Empirical Methods in Natural Language Processing}, pages 2529--2535, Copenhagen, Denmark. Association for Computational Linguistics.

\bibitem[{Marchisio et~al.(2024)Marchisio, Ko, B{\'e}rard, Dehaze, and Ruder}]{marchisio2024understanding}
Kelly Marchisio, Wei-Yin Ko, Alexandre B{\'e}rard, Th{\'e}o Dehaze, and Sebastian Ruder. 2024.
\newblock Understanding and mitigating language confusion in llms.
\newblock \emph{arXiv preprint arXiv:2406.20052}.

\bibitem[{McCann(2020)}]{mccann-2020-fugashi}
Paul McCann. 2020.
\newblock \href {https://www.aclweb.org/anthology/2020.nlposs-1.7} {fugashi, a tool for tokenizing {J}apanese in python}.
\newblock In \emph{Proceedings of Second Workshop for NLP Open Source Software (NLP-OSS)}, pages 44--51, Online. Association for Computational Linguistics.

\bibitem[{Morris et~al.(2023)Morris, Kuleshov, Shmatikov, and Rush}]{morris2023text}
John~X Morris, Volodymyr Kuleshov, Vitaly Shmatikov, and Alexander~M Rush. 2023.
\newblock Text embeddings reveal (almost) as much as text.
\newblock \emph{arXiv preprint arXiv:2310.06816}.

\bibitem[{Navigli and Ponzetto(2010)}]{navigli2010babelnet}
Roberto Navigli and Simone~Paolo Ponzetto. 2010.
\newblock Babelnet: Building a very large multilingual semantic network.
\newblock In \emph{Proceedings of the 48th annual meeting of the association for computational linguistics}, pages 216--225.

\bibitem[{{\"O}stling and Kurfal{\i}(2023)}]{ostling-kurfali-2023-language}
Robert {\"O}stling and Murathan Kurfal{\i}. 2023.
\newblock \href {https://doi.org/10.1162/coli_a_00491} {Language embeddings sometimes contain typological generalizations}.
\newblock \emph{Computational Linguistics}, 49(4):1003--1051.

\bibitem[{{\"O}stling and Tiedemann(2017)}]{ostling-tiedemann-2017-continuous}
Robert {\"O}stling and J{\"o}rg Tiedemann. 2017.
\newblock \href {https://aclanthology.org/E17-2102} {Continuous multilinguality with language vectors}.
\newblock In \emph{Proceedings of the 15th Conference of the {E}uropean Chapter of the Association for Computational Linguistics: Volume 2, Short Papers}, pages 644--649, Valencia, Spain. Association for Computational Linguistics.

\bibitem[{Pfeiffer et~al.(2023)Pfeiffer, Piccinno, Nicosia, Wang, Reid, and Ruder}]{pfeiffer-etal-2023-mmt5}
Jonas Pfeiffer, Francesco Piccinno, Massimo Nicosia, Xinyi Wang, Machel Reid, and Sebastian Ruder. 2023.
\newblock \href {https://doi.org/10.18653/v1/2023.findings-emnlp.132} {mm{T}5: Modular multilingual pre-training solves source language hallucinations}.
\newblock In \emph{Findings of the Association for Computational Linguistics: EMNLP 2023}, pages 1978--2008, Singapore. Association for Computational Linguistics.

\bibitem[{Pires et~al.(2019)Pires, Schlinger, and Garrette}]{pires-etal-2019-multilingual}
Telmo Pires, Eva Schlinger, and Dan Garrette. 2019.
\newblock \href {https://doi.org/10.18653/v1/P19-1493} {How multilingual is multilingual {BERT}?}
\newblock In \emph{Proceedings of the 57th Annual Meeting of the Association for Computational Linguistics}, pages 4996--5001, Florence, Italy. Association for Computational Linguistics.

\bibitem[{Ploeger et~al.(2024)Ploeger, Poelman, de~Lhoneux, and Bjerva}]{ploeger2024typological}
Esther Ploeger, Wessel Poelman, Miryam de~Lhoneux, and Johannes Bjerva. 2024.
\newblock \href {https://doi.org/10.48550/arXiv.2402.04222} {What is ``typological diversity'' in nlp?}
\newblock In \emph{Proceedings of the 2024 Conference on Empirical Methods in Natural Language Processing}.

\bibitem[{Poppi et~al.(2024)Poppi, Yong, He, Chern, Zhao, Yang, and Chi}]{poppi2024understandingfragilitymultilingualllms}
Samuele Poppi, Zheng-Xin Yong, Yifei He, Bobbie Chern, Han Zhao, Aobo Yang, and Jianfeng Chi. 2024.
\newblock \href {https://arxiv.org/abs/2410.18210} {Towards understanding the fragility of multilingual llms against fine-tuning attacks}.
\newblock \emph{Preprint}, arXiv:2410.18210.

\bibitem[{Post(2018)}]{post2018call}
Matt Post. 2018.
\newblock A call for clarity in reporting bleu scores.
\newblock \emph{arXiv preprint arXiv:1804.08771}.

\bibitem[{Rama and Kolachina(2012)}]{rama-kolachina-2012-good}
Taraka Rama and Prasanth Kolachina. 2012.
\newblock \href {https://aclanthology.org/C12-2095} {How good are typological distances for determining genealogical relationships among languages?}
\newblock In \emph{Proceedings of {COLING} 2012: Posters}, pages 975--984, Mumbai, India. The COLING 2012 Organizing Committee.

\bibitem[{Rzymski et~al.(2020)Rzymski, Tresoldi, Greenhill, Wu, Schweikhard, Koptjevskaja-Tamm, Gast, Bodt, Hantgan, Kaiping et~al.}]{rzymski2020database}
Christoph Rzymski, Tiago Tresoldi, Simon~J Greenhill, Mei-Shin Wu, Nathanael~E Schweikhard, Maria Koptjevskaja-Tamm, Volker Gast, Timotheus~A Bodt, Abbie Hantgan, Gereon~A Kaiping, et~al. 2020.
\newblock The database of cross-linguistic colexifications, reproducible analysis of cross-linguistic polysemies.
\newblock \emph{Scientific data}, 7(1):13.

\bibitem[{Sennrich et~al.(2024)Sennrich, Vamvas, and Mohammadshahi}]{sennrich-etal-2024-mitigating}
Rico Sennrich, Jannis Vamvas, and Alireza Mohammadshahi. 2024.
\newblock \href {https://aclanthology.org/2024.eacl-short.4} {Mitigating hallucinations and off-target machine translation with source-contrastive and language-contrastive decoding}.
\newblock In \emph{Proceedings of the 18th Conference of the European Chapter of the Association for Computational Linguistics (Volume 2: Short Papers)}, pages 21--33, St. Julian{'}s, Malta. Association for Computational Linguistics.

\bibitem[{Skirg{\aa}rd et~al.(2023)Skirg{\aa}rd, Haynie, Blasi, Hammarstr{\"o}m, Collins, Latarche, Lesage, Weber, Witzlack-Makarevich, Passmore et~al.}]{skirgaard2023grambank}
Hedvig Skirg{\aa}rd, Hannah~J Haynie, Dami{\'a}n~E Blasi, Harald Hammarstr{\"o}m, Jeremy Collins, Jay~J Latarche, Jakob Lesage, Tobias Weber, Alena Witzlack-Makarevich, Sam Passmore, et~al. 2023.
\newblock Grambank reveals the importance of genealogical constraints on linguistic diversity and highlights the impact of language loss.
\newblock \emph{Science Advances}, 9(16):eadg6175.

\bibitem[{Song and Raghunathan(2020)}]{10.1145/3372297.3417270}
Congzheng Song and Ananth Raghunathan. 2020.
\newblock \href {https://doi.org/10.1145/3372297.3417270} {Information leakage in embedding models}.
\newblock In \emph{Proceedings of the 2020 ACM SIGSAC Conference on Computer and Communications Security}, CCS '20, page 377–390, New York, NY, USA. Association for Computing Machinery.

\bibitem[{Song et~al.(2024)Song, Huang, Zhou, and Ma}]{song2024multilingual}
Jiayang Song, Yuheng Huang, Zhehua Zhou, and Lei Ma. 2024.
\newblock Multilingual blending: Llm safety alignment evaluation with language mixture.
\newblock \emph{arXiv preprint arXiv:2407.07342}.

\bibitem[{Sun et~al.(2013)}]{jieba_chinese}
Andy Sun et~al. 2013.
\newblock jieba.
\newblock \url{https://github.com/fxsjy/jieba}.
\newblock Accessed: 2024-10-15.

\bibitem[{Talat et~al.(2022)Talat, N{\'e}v{\'e}ol, Biderman, Clinciu, Dey, Longpre, Luccioni, Masoud, Mitchell, Radev et~al.}]{talat2022you}
Zeerak Talat, Aur{\'e}lie N{\'e}v{\'e}ol, Stella Biderman, Miruna Clinciu, Manan Dey, Shayne Longpre, Sasha Luccioni, Maraim Masoud, Margaret Mitchell, Dragomir Radev, et~al. 2022.
\newblock You reap what you sow: On the challenges of bias evaluation under multilingual settings.
\newblock In \emph{Proceedings of BigScience Episode\# 5--Workshop on Challenges \& Perspectives in Creating Large Language Models}, pages 26--41.

\bibitem[{Touvron et~al.(2023)Touvron, Martin, Stone, Albert, Almahairi, Babaei, Bashlykov, Batra, Bhargava, Bhosale et~al.}]{touvron2023llama}
Hugo Touvron, Louis Martin, Kevin Stone, Peter Albert, Amjad Almahairi, Yasmine Babaei, Nikolay Bashlykov, Soumya Batra, Prajjwal Bhargava, Shruti Bhosale, et~al. 2023.
\newblock Llama 2: Open foundation and fine-tuned chat models.
\newblock \emph{arXiv preprint arXiv:2307.09288}.

\bibitem[{Vu et~al.(2022)Vu, Barua, Lester, Cer, Iyyer, and Constant}]{vu_overcoming_2022}
Tu~Vu, Aditya Barua, Brian Lester, Daniel Cer, Mohit Iyyer, and Noah Constant. 2022.
\newblock \href {https://doi.org/10.18653/v1/2022.emnlp-main.630} {Overcoming catastrophic forgetting in zero-shot cross-lingual generation}.
\newblock In \emph{Proceedings of the 2022 Conference on Empirical Methods in Natural Language Processing}, pages 9279--9300. Association for Computational Linguistics.

\bibitem[{Wang et~al.(2024)Wang, Xu, He, Rubinstein, and Cohn}]{wang_backdoor_2024}
Jun Wang, Qiongkai Xu, Xuanli He, Benjamin Rubinstein, and Trevor Cohn. 2024.
\newblock \href {https://doi.org/10.18653/v1/2024.naacl-long.254} {Backdoor attacks on multilingual machine translation}.
\newblock In \emph{Proceedings of the 2024 Conference of the North American Chapter of the Association for Computational Linguistics: Human Language Technologies (Volume 1: Long Papers)}, pages 4515--4534. Association for Computational Linguistics.

\bibitem[{Wichmann et~al.(2012)Wichmann, M{\"u}ller, Velupillai, Wett, Brown, Molochieva, Bishoffberger, Holman, Sauppe, Brown et~al.}]{wichmann2012automated}
S{\o}ren Wichmann, Andr{\'e} M{\"u}ller, Viveka Velupillai, Annkathrin Wett, Cecil~H Brown, Zarina Molochieva, Julia Bishoffberger, Eric~W Holman, Sebastian Sauppe, Pamela Brown, et~al. 2012.
\newblock The automated similarity judgment program (asjp): the asjp database (version 15).

\bibitem[{Xue(2020)}]{xue2020mt5}
L~Xue. 2020.
\newblock mt5: A massively multilingual pre-trained text-to-text transformer.
\newblock \emph{arXiv preprint arXiv:2010.11934}.

\bibitem[{Yong et~al.(2024)Yong, Menghini, and Bach}]{yong_low-resource_2024}
Zheng-Xin Yong, Cristina Menghini, and Stephen~H. Bach. 2024.
\newblock \href {https://arxiv.org/abs/2310.02446 [cs]} {Low-resource languages jailbreak {GPT}-4}.
\newblock \emph{Preprint}, arxiv:2310.02446 [cs].

\bibitem[{Zhang et~al.(2020)Zhang, Williams, Titov, and Sennrich}]{zhang-etal-2020-improving}
Biao Zhang, Philip Williams, Ivan Titov, and Rico Sennrich. 2020.
\newblock \href {https://doi.org/10.18653/v1/2020.acl-main.148} {Improving massively multilingual neural machine translation and zero-shot translation}.
\newblock In \emph{Proceedings of the 58th Annual Meeting of the Association for Computational Linguistics}, pages 1628--1639, Online. Association for Computational Linguistics.

\bibitem[{Zhu et~al.(2024)Zhu, Liu, Dong, Xu, Huang, Kong, Chen, and Li}]{zhu2024multilingual}
Wenhao Zhu, Hongyi Liu, Qingxiu Dong, Jingjing Xu, Shujian Huang, Lingpeng Kong, Jiajun Chen, and Lei Li. 2024.
\newblock Multilingual machine translation with large language models: Empirical results and analysis.
\newblock In \emph{Findings of the Association for Computational Linguistics: NAACL 2024}, pages 2765--2781.

\bibitem[{Zou et~al.(2023)Zou, Wang, Carlini, Nasr, Kolter, and Fredrikson}]{gcg}
Andy Zou, Zifan Wang, Nicholas Carlini, Milad Nasr, J.~Zico Kolter, and Matt Fredrikson. 2023.
\newblock Universal and transferable adversarial attacks on aligned language models.
\newblock \emph{arXiv preprint arXiv:2307.15043}.

\end{thebibliography}

\appendix

\section{Datasets for Language Confusion}\label{sec:datasets}
\renewcommand{\arraystretch}{1.5} 

\begin{table*}[]
    \centering
     \resizebox{\linewidth}{!}{
    \begin{tabular}{c|ccccp{5cm}c}
    \toprule
         & Dataset name  & Nature of Data & $|L|$ & $|D|$ & Languages & $W$  \\
         \midrule
 \multirow{4}{*}{\rotatebox{90}{\textbf{Monolingual}}}
  & Aya~\citep{iyer-etal-2024-exploring} & Human-generated & 100 & 500 &eng, tur, arb, cmn, por & 9\\
  & Dolly~\citep{iyer-etal-2024-exploring} & MT post-edited & 100 & 500 & hin, rus, fra, arb, esp & 10 \\
  & Okapi~\citep{lai-etal-2023-okapi} & Synthetic+MT & 100 & 1.2k & eng, fra, ita, deu, cmn, vie, rus, esp, find, por, arb, hin, esp, fra, jap, kor & 13 \\
  & Native prompts~\citep{marchisio2024understanding} 
  & Human-generated & 100 & 500 & esp, fra, jap, kor & 19\\
  \midrule 
 \multirow{3}{*}{\rotatebox{90}{\makecell{{\textbf{Crosslingual}}}}}
  & Okapi~\citep{lai-etal-2023-okapi} & Synthetic & 100 & 1.5k & $\mathcal{L}$ & 15 \\

   & ShareGPT~(\url{https://sharegpt.com/}) & Human-generated & 100 & 1.5k & $\mathcal{L}$ & 18 \\

  & Complex prompts~\citep{marchisio2024understanding} & Human-generated & 99 & 1.5k & $\mathcal{L}$ & 159 \\
         \bottomrule
    \end{tabular}
    }
    \caption{Data Sources in the \textbf{LCB} for monolingual and crosslingual generation~\citep{marchisio2024understanding}. $|D|$ is the total number of examples per data source and $|L|$ is the number of examples per language. For the crosslingual setting, the model is instructed in English to generate in the target language $l \in \mathcal{L}$ where $\mathcal{L}$={fra, deu, esp, por, ita, jap, kor, cmn, arb, tur, hin, rus, ind, vie}. $W$ is the median length in words of the prompts in each dataset.}
    \label{tab:lcb_datasources}
\end{table*}
\renewcommand{\arraystretch}{1} 

\paragraph{Language Confusion Benchmark}
~\citet{marchisio2024understanding} create and release a language confusion benchmark covering 15 languages, sourcing prompts from publicly available multilingual instruction datasets, and also creating new data with more complex promts, see detailed data sources in Table~\ref{tab:lcb_datasources}.

Additionally, the binary metrics such as Line Pass Rate (LPR) and Word Pass Rate (WPR) are defined in~\citet{marchisio2024understanding} to measure whether a response contains any instance of a) a line in an incorrect language and b) an isolated English word/phrase for languages using non-Latin scripts. 

\textbf{LPR} calculates the percentage of model responses that pass the line-level language confusion detector without error. A response is ``correct'' if all lines match the user's desired language. 

\begin{equation}
    \begin{aligned}
   LPR = \frac{|R \backslash E_L|}{|R|}
    \end{aligned}
\end{equation}
where $R$ is the set of all responses and $E_L$ is the set of responses that contain line-level errors.

\textbf{WPR} measures the percentage of responses where all words are in the desired language. 
\begin{equation}
    \begin{aligned}
   WPR = \frac{|(R \backslash E_L)|\backslash E_W}{|R\backslash E_L|}
    \end{aligned}
\end{equation}
where $R$ is the set of all responses,
$E_L$ is the set of responses with line-level errors, 
and $E_W$ the set of responses with word-level errors.

We reproduce the \textbf{LPR} and \textbf{WPR} on \textbf{LCB} for both crosslingual and monolingual settings (as shown in Table~\ref{tab:lcb_lpr_reproduction} and~\ref{tab:lcb_wpr_reproduction}).
The detailed results applying \textit{language confusion entropy} to \textbf{LCB} are presented in Table~\ref{tab:lcb_confusion_entropy_llms}, for comparison.

\begin{table*}[ht!]
    \centering
     \resizebox{\linewidth}{!}{
    \begin{tabular}{l|cccl}
    \toprule
        \textbf{LLM}  & \textbf{Transformer} & \textbf{\#Languages} &  \textbf{ Parameters} & \textbf{Reference} \\
        \midrule
        \underline{\textbf{LCB}} & & & &   \\
       Command R & Decoder-only& - & 35B &  \url{https://cohere.com/blog/command-r}\\
       Command R+ & Decoder-only & - & 104B & \url{https://cohere.com/blog/command-r-plus-microsoft-azure}\\
       GPT-3.5 Turbo &Decoder-only &  - &  - & ~\citet{brown2020language}  \\
        GPT-4 Turbo & Decoder-only & - &  - & ~\citet{achiam2023gpt}  \\
        Mistral Large & Decoder-only & -  &  - & \url{https://mistral.ai/news/mistral-large/}\\
        Mistral 8x7B  & Decoder-only & -  &  7B & ~\citet{jiang2024mixtral}, \\
        Llama 2 70B Instruct & Decoder-only & - &  70B & ~\citet{touvron2023llama}  \\
         Llama 3 70B Instruct & Decoder-only & - & 70B & \url{https://ai.meta.com/blog/meta-llama-3/} \\
        \midrule
         \underline{\textbf{MTEI}} & & & &   \\
         mT5-base & Encoder-Decoder & 102 & 580M  & ~\citet{xue2020mt5} \\
         multilingual-e5-base & Encoder & 94 & 580M &  \url{https://huggingface.co/intfloat/multilingual-e5-base} \\  
         \bottomrule
    \end{tabular}}
    \caption{The Evaluated LLMs.}
    \label{tab:llms}
\end{table*}

\paragraph{Multilingual Textual Embedding Inversion}
Textual embedding inversion has presented a standing challenge in LLM security, where the private texts can be reconstructed from evasdroped embeddings from Embeddings as a Service (EaaS), by training an attacker model based on the embeddings extracted from the black-box embedders~\citep{10.1145/3372297.3417270, Lyu2020DifferentiallyPR, kim-etal-2022-toward, morris2023text, chen-etal-2024-text,chen2024oddsovercomingtypologyscript}. However, most work was done in monolingual settings, mostly in English, other than the recent work expands the language space to four Romance and Germanic languages in Latin script ~\citep{chen-etal-2024-text} and in \citet{chen2024oddsovercomingtypologyscript}, the inversion atacks are extended to 20 languages across 8 families and 12 scripts (see Table~\ref{tab:inversion-languages}). The trained inversion attack model consists of a \textbf{base} model and a \textbf{corrector} model, where a base model is a text-to-text generation model, while a corrector model is used to bring closer the generated embeddings and attacked embeddings in the embedding space. While in the evaluation phase, three stages are reported: \textbf{base}, \textbf{step1 (corrector model)} and \textbf{step50+sbeam8 (corrector model with beam search with sequence length 8)}. The inversion model is trained with \textbf{mT5}~\citep{xue2020mt5} as base model and multilingual-e5-base~\footnote{Huggingface: intfloat/multilingual-e5-base} as black-box encoder. The samples of the curated dataset are shown in Table~\ref{tab:examples_mtei}.

We apply \textit{language confusion entropy} to \textbf{eval} and \textbf{train} languages in the monolingual and crosslingual settings at both line and word levels, while comparing the BLEU score in the regarding scenario (ref. Table~\ref{tab:mtei_lc_inversion_eval} and~\ref{tab:mtei_lc_bleu_train}).

\begin{table}[ht!]
    \centering
     \resizebox{0.7\linewidth}{!}{ 
    \begin{tabular}{c|c|cccc}
    \toprule 
        & & \textbf{$H_{\mathbf{C}}$[L]} &   \textbf{$H_{\mathbf{C}}$[W]} & BLEU   \\
        \midrule
                \multirow{3}{*}{\textbf{All}} &
         \textbf{$H_{\mathbf{C}}$[W]} & $0.93$*** &  & \\
       & \textbf{BLEU} & $0.66$***	& $0.71$***& \\
      &  \textbf{mT5} & $0.32$***	& $0.24$ &	$0.09$	\\
   
        \midrule
        
        \multirow{3}{*}{\textbf{Monolingual}} &
         \textbf{$H_{\mathbf{C}}$[W]} & $0.63$*** &  & \\
       &\textbf{ BLEU} & $0.37$*** &	$0.33$** & \\
      &  \textbf{mT5} & 0.05&	$0.42$* &	$0.09$	\\

         \midrule
   \multirow{3}{*}{\textbf{Crosslingual}} &
        \textbf{$H_{\mathbf{C}}$[W]} & $0.93$*** &  & \\
       & \textbf{BLEU} & $0.66$***&	$0.71$*** & \\
      &  \textbf{mT5} & $0.32$&	$0.23$&	$0.09$	\\

        \bottomrule
    \end{tabular}
    }
 \caption{Spearman Correlations among  $H_{\mathbf{C}}$  at Line Level and Word Level and BLEU score for Multilingual Textual Embedding Inversion, and the percentage of pre-training data in mT5, for train languages.}  
\label{tab:inversion_entropy_bleu_train_corr}
\end{table}

\begin{table*}[!h]
    \centering
     \resizebox{\linewidth}{!}{
    \begin{tabular}{ccccccccc}
    \toprule
        \textbf{Language} & \textbf{ISO 639} & \textbf{Lang. Family} & \textbf{Lang. Script} & \textbf{Script ISO} & \textbf{Directionality} & \textbf{\#Samples(Train)} & \textbf{WO} & \textbf{mT5 (\%)}  \\ 
        \midrule
        Arabic & arb & Semitic & Arabic & Arab & RTL & 1M & VSO & 1.66 \\ 
        Urdu & urd & Indo-Aryan & Arabic & Arab & RTL & 600K & SOV  & 0.61\\ 
        Kazakh & kaz & Turkic & Cyrillic & Cyrl & LTR & 1M & SOV  & 0.65\\ 
        Mongolian & mon & Mongolic & Cyrillic & Cyrl & LTR & 1M  & SOV &  0.62\\ 
        Hindi & hin & Indo-Aryan & Devanagari & Deva & LTR  &  600K & SOV & 1.21\\ 
        Gujarati & guj & Indo-Aryan & Gujarati & Gujr & LTR &  600K & SOV & 0.43 \\ 
        Punjabi & pan & Indo-Aryan & Gurmukhi & Guru & LTR &  600K & SOV & 0.37 \\ 
        Chinese & cmn & Sino-Tibetan & Haqniqdoq & Hani & LTR & 1M & SVO & 1.67 \\ 
        Hebrew & heb & Semitic & Hebrew & Hebrewr & RTL & 1M & SVO & 1.06 \\ 
        Japanese & jpn & Japonic & Japanese & Jpan & LTR & 1M & SOV& 1.92\\ 
        German & deu & Germanic & Latin & Latn & LTR & 1M  & Non-Dominant & 3.05\\ 
        Turkish & tur & Turkic & Latin & Latn & LTR & 1M & SOV& 1.93 \\ 
        Amharic & amh & Semitic & Ethiopian & Ethi & LTR &  - & SOV & 0.29\\ 
        Sinhala & sin & Indo-Aryan & Sinhala & Sinh & LTR  & -& SOV& 0.41\\ 
        Korean & kor & Koreanic & Hangul & Hang & LTR & - & SOV & 1.14 \\ 
        Finnish & fin & Uralic & Latin & Latn & LTR & - & SVO & -\\ 
        Hungarian & hun & Uralic & Latin & Latn & LTR & -  & Non-Dominant& 1.48 \\ 
        Yiddish & ydd & Germanic & Hebrew & Hebrewr & RTL & - & SVO &  0.28\\ 
        Maltese & mlt & Semitic & Latin & Latn & LTR & - & Non-Dominant & 0.64 \\ 
        Meadow Mari & mhr & Uralic & Cyrillic & Cyrl & LTR & - & SOV& - \\

        \bottomrule
    \end{tabular}
    }
    \caption{Languages and their Language Characteristics, i.e., Language Family, Language Script, Directionality of the Script, Number of Training Samples for Inversion Models, Word Order of Subject, Object and Verb in Multilingual Inversion Attack~\citep{chen2024oddsovercomingtypologyscript} and the Percentage of the language in Pre-training data in mT5~\citep{xue2020mt5}.}
    \label{tab:inversion-languages}
\end{table*}

\begin{table*}[!ht]
    \centering
    \resizebox{\linewidth}{!}{
    \begin{tabular}{lllll}
    \toprule
    
        \textbf{Model} & $\mathcal{L}_{\text{train}}$ & $\mathcal{L}_{\text{eval}}$ & \textbf{Eval Step} & \textbf{Predicted Language Distribution}  \\ 
        \midrule
                \textsc{me5} & Hindi & German & Base & \{eng: 0.27, deu: 0.34, hin: 0.24, fra: 0.01, nld: 0.01, fin: 0.01, mar: 0.02, nep: 0.01\} \\ 

        \textsc{me5} & Hindi & German & Step1 & \{eng: 0.17, deu: 0.47, hin: 0.24, mar: 0.02, fra: 0.01, nep: 0.01, nld: 0.01\} \\ 
        \textsc{me5} & Hindi & German & Step50+sbeam8 & \{hin: 0.38, mar: 0.04, deu: 0.37, eng: 0.15, fra: 0.01\} \\ 
                \textsc{me5} & Hindi & Yiddish & Base & \{mar: 0.12, hin: 0.83, eng: 0.02, deu: 0.01, nep: 0.01\} \\ 

        \textsc{me5} & Hindi & Yiddish & Step1 & \{hin: 0.82, eng: 0.02, mar: 0.13, deu: 0.01\} \\ 
        \textsc{me5} & Hindi & Yiddish & Step50+sbeam8 & \{hin: 0.81, mar: 0.12, deu: 0.01, eng: 0.04\} \\ 
                \textsc{me5} & Hindi & Hebrew & Base & \{hin: 0.92, eng: 0.03, mar: 0.03\} \\ 

        \textsc{me5} & Hindi & Hebrew & Step1 & \{hin: 0.94, mar: 0.03, eng: 0.02\} \\ 
        \textsc{me5} & Hindi & Hebrew & Step50+sbeam8 & \{hin: 0.94, eng: 0.02, mar: 0.03\} \\ 
                \textsc{me5} & Hindi & Arabic & Base & \{eng: 0.63, hin: 0.32, mar: 0.02, nep: 0.01, fra: 0.01\} \\ 

        \textsc{me5} & Hindi & Arabic & Step1 & \{eng: 0.61, hin: 0.33, mar: 0.03\} \\ 
        \textsc{me5} & Hindi & Arabic & Step50+sbeam8 & \{eng: 0.62, hin: 0.31, mar: 0.04\} \\ 
        \bottomrule
    \end{tabular}}
    \caption{Examples of Dataset MTEI. The probabilities for \textit{unidentified} languages are omitted. }
    \label{tab:examples_mtei}
\end{table*}

\begin{table*}[!ht]
\centering
     \resizebox{\linewidth}{!}{ 
     \begin{tabular}{l|>{\columncolor{gray!20}}ccccccccccccccccc}
     \toprule
    
      & \textbf{AVG} & \textbf{French} & \textbf{Spanish} & \textbf{Italian} & \textbf{German} & \textbf{Russian} & \textbf{Portuguese} & \textbf{Turkish} & \textbf{Vietnamese} & \textbf{Chinese} & \textbf{Korean} & \textbf{Arabic} & \textbf{Japanese} & \textbf{Hindi} & \textbf{Indonesian} \\ 
      
        \midrule
         \textit{ \textbf{\underline{Monolingual (Line)} } }& \\
        \textbf{Command R} & \textbf{0.0306} & 0.0272 & 0.0381 & 0.0092 & 0.0430 & 0.0093 & 0.0195 & 0.0042 & 0.0099 & 0 & 0 & 0.0017 & 0.0171 & \textbf{0.0111} & 0.2384 \\ 
        \textbf{Command R+} & 0.0355 & \textbf{0.0127} & 0.0325 & 0 & 0.0373 & 0.0309 & 0.0219 & 0.0100 & 0.0264 & 0 & 0 & 0.0034 & 0.0158 & 0.0387 & 0.2673 \\ 
        \textbf{GPT-3.5 Turbo} & 0.0317 & 0.0250 & \textbf{0.0296} & 0.0103 & 0.0276 & 0.0168 & 0.0292 & 0.0050 & 0.0070 & 0 & 0 & 0 & \textbf{0.0057} & 0.0546 & 0.2337 \\ 
        \textbf{GPT-4 Turbo} & 0.0381 & 0.0508 & 0.0468 & 0.0051 & 0.0316 & 0.0636 & 0.0228 & 0 & 0.0033 & 0.0079 & 0 & 0 & 0.0083 & 0.0639 & 0.2289 \\ 
         \textbf{Mistral 8x7B} & 0.0766 & 0.0399 & 0.0510 & 0.0645 & 0.0229 & 0.0414 & 0.0784 & 0.0252 & 0.1037 & 0.1116 & 0.0787 & 0.0583 & 0.1394 & 0.0540 & 0.2036 \\ 
        \textbf{Mistral Large} & 0.0799 & 0.0145 & 0.0351 & 0.0202 & 0.0139 & 0.0156 & 0.0643 & 0.0378 & 0.2245 & 0.1011 & 0.0329 & 0.1231 & 0.0756 & 0.0663 & 0.2936 \\ 

        \textbf{Llama 2 70B-I} & 0.1266 & 0.0729 & 0.0548 & 0.1500 & 0.1438 & 0.0704 & 0.0569 & 0.1347 & 0.2501 & 0.1113 & 0.0501 & 0.1713 & 0.0784 & 0.1279 & 0.2999 \\ 
        \textbf{Llama 3 70B-I} & 0.1197 & 0.0483 & 0.0557 & 0.0300 & 0.0556 & 0.0955 & 0.0654 & 0.2009 & 0.1416 & 0.1518 & 0.1720 & 0.1879 & 0.1579 & 0.1720 & \textbf{0.1412} \\ 
       
        \midrule        
        \textit{\textbf{\underline{Crosslingual (Line)}}} & \\ 
        \textbf{Command R} & 0.1555 & 0.1942 & 0.1651 & 0.1723 & 0.1928 & 0.1357 & 0.1458 & 0.1441 & 0.1579 & 0.1007 & 0.1169 & 0.1159 & 0.1522 & 0.1395 & \textbf{0.2437} \\ 
        \textbf{Command R+} & 0.1216 & 0.1234 & 0.1635 & 0.0810 & 0.1040 & 0.1034 & 0.1555 & 0.0817 & 0.1204 & 0.0610 & 0.0617 & \textbf{0.0534} & 0.0691 & 0.1424 & 0.3825 \\ 
        \textbf{GPT-3.5 Turbo} & 0.1025 & 0.1219 & 0.1085 & 0.1042 & 0.1119 & \textbf{0.0749} & 0.1169 & 0.0945 & 0.0772 & \textbf{0.0527} & \textbf{0.0596} & 0.0764 & \textbf{0.0624} & \textbf{0.1163} & 0.2584 \\ 
        \textbf{GPT-4 Turbo} & \textbf{0.0987} & \textbf{0.0721} & \textbf{0.0956} & \textbf{0.1002} & \textbf{0.0778} & 0.0876 & \textbf{0.0909} & \textbf{0.0733} & \textbf{0.0742} & 0.0665 & 0.0600 & 0.0666 & 0.0981 & 0.1612 & 0.2577 \\ 
         \textbf{Mistral 8x7B} & 0.1675 & 0.1341 & 0.1332 & 0.1242 & 0.1582 & 0.1304 & 0.1288 & 0.1156 & 0.1458 & 0.1604 & 0.1743 & 0.1506 & 0.1977 & 0.3053 & 0.2861 \\ 
        \textbf{Mistral Large} & 0.2274 & 0.1341 & 0.2122 & 0.1808 & 0.1441 & 0.1496 & 0.2509 & 0.1410 & 0.1817 & 0.2944 & 0.3052 & 0.2816 & 0.3416 & 0.2407 & 0.3262 \\ 
        \textbf{Llama 2 70B-I} & 0.2649 & 0.1517 & 0.1339 & 0.2330 & 0.2657 & 0.2186 & 0.2022 & 0.2280 & 0.2532 & 0.2817 & 0.3574 & 0.3719 & 0.3233 & 0.3443 & 0.3433 \\ 
        \textbf{Llama 3 70B-I} & 0.4631 & 0.3906 & 0.3192 & 0.4324 & 0.4226 & 0.4495 & 0.3503 & 0.5696 & 0.5360 & 0.3872 & 0.3777 & 0.5200 & 0.4822 & 0.5531 & 0.6935 \\ 
       
        \midrule
        
        \textit{\textbf{\underline{Monolingual (Word)} }} &\\ 
        \textbf{Command R} & 0.5563 & 0.9016 & 0.9424 & 0.9513 & 0.4098 & 0.6312 & 1.0095 & 0.5445 & 0.4059 & 0.1974 & \textbf{0.0031} & 0.1424 & 0.0462 & 0.4161 & 1.1865 \\ 
        \textbf{Command R+} & 0.5522 & 0.9034 & 0.9687 & 0.9363 & \textbf{0.3873} & 0.6478 & 1.0185 & 0.5003 & 0.3812 & 0.1232 & 0.0057 & 0.1327 & 0.0785 & 0.4007 & 1.2458 \\ 
        \textbf{GPT-3.5 Turbo} & \textbf{0.5344} & 0.9031 & 0.9348 & 0.9567 & 0.4325 & 0.6322 & \textbf{0.9457} & \textbf{0.4281} & 0.3587 & \textbf{0.0913} & 0.0088 & \textbf{0.1053} & 0.0370 & \textbf{0.3937} & 1.2532 \\ 
        \textbf{GPT-4 Turbo} & 0.5426 & \textbf{0.8656} & 0.9456 & 0.9284 & 0.4241 & 0.6206 & 0.9818 & 0.4932 & 0.3217 & 0.1287 & 0.0146 & 0.1355 & \textbf{0.0307} & 0.4232 & 1.2823 \\ 
        
          \textbf{Mistral 8x7B} & 0.7431 & 0.9763 & 0.9157 & 1.0027 & 0.4477 & 0.6947 & 0.9746 & 0.6763 & 0.4134 & 0.8854 & 0.4960 & 0.4330 & 0.8628 & 0.6106 & 1.0134 \\ 
        \textbf{Mistral Large} & 0.5703 & 0.8829 & \textbf{0.9099} & \textbf{0.9083} & 0.4270 & \textbf{0.6191} & 0.9827 & 0.6473 & \textbf{0.3199} & 0.2191 & 0.2568 & 0.3137 & 0.2484 & 0.4472 & \textbf{0.8016} \\ 
        
        \textbf{Llama 2 70B-I} & 0.8022 & 1.0541 & 0.9257 & 1.0109 & 0.8092 & 0.6939 & 1.0000 & 1.2333 & 0.5748 & 0.5989 & 0.3960 & 0.5578 & 0.5229 & 0.5676 & 1.2862 \\ 
        \textbf{Llama 3 70B-I} & 0.8657 & 1.0519 & 0.9427 & 1.0139 & 0.8992 & 0.6325 & 1.0028 & 1.5459 & 0.4827 & 0.8949 & 0.8120 & 0.6904 & 0.6017 & 0.6831 & 0.8659 \\ 
      
        \midrule
        
       \textit{ \textbf{\underline{Crosslingual (Word)}} }&  \\ 
        \textbf{Command R} & 0.6796 & 1.0941 & 1.0038 & 1.0896 & 0.6588 & 0.6618 & \textbf{0.9487} & 0.5579 & 0.4664 & 0.3347 & 0.1988 & 0.3190 & 0.4943 & 0.4836 & \textbf{1.2030} \\ 
        \textbf{Command R+} & 0.6049 & \textbf{0.9606} & \textbf{0.9380} & \textbf{0.9444} & \textbf{0.4382} & 0.6684 & 1.0785 & \textbf{0.5316} & 0.4572 & 0.2081 & \textbf{0.0670} & \textbf{0.1616} & 0.1870 & 0.5074 & 1.3212 \\ 
        \textbf{GPT-3.5 Turbo} & \textbf{0.5951} & 1.0295 & 0.9439 & 0.9894 & 0.5858 & \textbf{0.5796} & 1.0159 & 0.5327 & \textbf{0.3659} & \textbf{0.1273} & 0.0933 & 0.1727 & \textbf{0.1850} & \textbf{0.4478} & 1.2626 \\ 
        \textbf{GPT-4 Turbo} & 0.6348 & 1.0025 & 0.9619 & 0.9890 & 0.5257 & 0.6425 & 1.0182 & 0.5568 & 0.4089 & 0.2542 & 0.1182 & 0.2552 & 0.3012 & 0.5073 & 1.3451 \\ 
     
        \textbf{Mistral 8x7B} & 0.8556 & 1.0509 & 1.0205 & 1.0143 & 0.6203 & 0.7656 & 1.0186 & 0.6174 & 0.5099 & 0.7446 & 0.6496 & 0.5800 & 0.9520 & 1.1017 & 1.3330 \\ 
        \textbf{Mistral Large} & 0.9647 & 1.2351 & 1.1284 & 1.1409 & 0.6319 & 0.7850 & 1.1244 & 0.6909 & 0.6404 & 1.1388 & 0.7054 & 0.7411 & 1.2626 & 0.8266 & 1.4551 \\ 
           \textbf{Llama 2 70B-I} & 0.9676 & 1.2830 & 0.9488 & 1.1583 & 0.9822 & 0.7885 & 1.0626 & 0.6962 & 0.6398 & 1.1089 & 0.8069 & 0.6929 & 1.1530 & 0.9268 & 1.2983 \\ 
        \textbf{Llama 3 70B-I} & 0.9757 & 1.2111 & 1.1366 & 1.3193 & 0.8329 & 1.0911 & 1.2465 & 0.9568 & 0.7462 & 0.7249 & 0.3435 & 0.8228 & 0.8275 & 0.8733 & 1.5274 \\ 
        \bottomrule
    \end{tabular}}
    \caption{Language Confusion measured by Language Confusion Entropy for LCB for monolingual and crosslingual settings at line and word level for each Language for LLMs. }
    \label{tab:lcb_confusion_entropy_llms}
\end{table*}

\begin{table*}[!ht]
    \centering

      \resizebox{\linewidth}{!}{ 
     \begin{tabular}{l|>{\columncolor{gray!20}}ccccccccccccccccc}
    \toprule
        \textbf{LPR} & \textbf{AVG} & \textbf{French} & \textbf{Spanish} & \textbf{Italian} & \textbf{German} & \textbf{Russian} & \textbf{Portuguese} & \textbf{Turkish} & \textbf{Vietnamese} & \textbf{Chinese} & \textbf{Korean} & \textbf{Arabic} & \textbf{Japanese} & \textbf{Hindi} & \textbf{Indonesian} \\ 
         \midrule
         \textit{\textbf{\underline{Monolingual}} }  & \\
        \textbf{Command R} & 98.50 & 99.33 & 95.67 & 99.00 & 98.00 & 100.00 & 98.50 & 99.00 & 99.00 & 98.50 & 100.00 & 100.00 & 100.00 & 100.00 & 92.00 \\ 
        \textbf{Command R+} & 99.19 & 99.67 & 99.33 & 100.00 & 100.00 & 100.00 & 97.50 & 100.00 & 99.00 & 97.50 & 100.00 & 99.67 & 99.00 & 100.00 & 97.00 \\ 
        \textbf{GPT-3.5 Turbo} & 99.05 & 100.00 & 99.67 & 100.00 & 100.00 & 100.00 & 98.00 & 100.00 & 99.00 & 97.00 & 100.00 & 100.00 & 98.00 & 99.00 & 96.00 \\ 
        \textbf{GPT-4 Turbo} & \textbf{99.26} & 99.33 & 99.33 & 99.00 & 100.00 & 100.00 & 98.00 & 100.00 & 100.00 & 99.00 & 100.00 & 99.00 & 100.00 & 100.00 & 96.00 \\ 
        \textbf{Mistral 8x7B} & 71.08 & 95.33 & 89.33 & 72.00 & 91.00 & 65.00 & 85.00 & 90.00 & 57.00 & 45.50 & 61.00 & 48.00 & 67.00 & 71.00 & 58.00 \\
        \textbf{Mistral Large} & 67.82 & 100.00 & 99.00 & 99.00 & 98.00 & 98.00 & 79.50 & 71.00 & 29.00 & 66.00 & 64.00 & 48.00 & 48.00 & 19.00 & 31.00 \\ 

        \textbf{Llama 2 70B-I} & 44.65 & 87.67 & 95.67 & 72.00 & 59.00 & 89.00 & 91.00 & 33.00 & 17.00 & 10.50 & 0.00 & 0.33 & 7.00 & 1.00 & 62.00 \\ 
        \textbf{Llama 3 70B-I} & 42.15 & 88.67 & 98.33 & 88.00 & 31.00 & 77.00 & 95.50 & 18.00 & 10.00 & 8.00 & 0.00 & 21.67 & 10.00 & 23.00 & 21.00 \\ 

           \midrule
         \textit{\textbf{\underline{Crosslingual}} } & \\
        
        \textbf{Command R} & 77.84 & 86.83 & 84.17 & 74.00 & 72.00 & 77.00 & 79.80 & 75.50 & 74.00 & 84.40 & 77.00 & 80.83 & 74.00 & 74.25 & 76.00 \\ 
        \textbf{Command R+} & \textbf{93.77} & 95.50 & 95.50 & 95.25 & 93.75 & 94.75 & 92.00 & 94.00 & 93.00 & 92.80 & 93.25 & 96.50 & 95.00 & 92.75 & 88.75 \\ 
        \textbf{GPT-3.5 Turbo} & 92.83 & 94.00 & 96.50 & 93.50 & 92.75 & 93.75 & 93.20 & 92.00 & 93.50 & 90.60 & 92.75 & 95.33 & 90.75 & 93.75 & 87.25 \\ 
        \textbf{GPT-4 Turbo} & 93.13 & 95.00 & 96.17 & 93.50 & 94.75 & 92.50 & 93.80 & 93.50 & 92.50 & 92.40 & 92.25 & 94.00 & 90.75 & 93.25 & 89.50 \\ 
         
        \textbf{Mistral 8x7B} & 69.73 & 87.17 & 84.17 & 81.75 & 80.00 & 70.50 & 81.60 & 79.50 & 71.00 & 52.60 & 58.00 & 53.50 & 60.25 & 47.25 & 69.00 \\ 
        \textbf{Mistral Large} & 62.26 & 86.00 & 83.83 & 74.25 & 80.50 & 72.00 & 70.60 & 67.25 & 48.25 & 54.20 & 46.75 & 42.00 & 45.25 & 48.50 & 52.25 \\
        \textbf{Llama 2 70B-I} & 40.53 & 79.33 & 86.50 & 67.75 & 54.00 & 51.00 & 82.00 & 26.25 & 19.55 & 10.84 & 3.58 & 6.33 & 14.00 & 16.25 & 50.00 \\ 
        \textbf{Llama 3 70B-I} & 34.81 & 70.83 & 79.67 & 49.25 & 33.75 & 48.00 & 70.80 & 17.53 & 16.53 & 5.80 & 0.58 & 26.33 & 3.53 & 40.50 & 24.25 \\ 
        \bottomrule
    \end{tabular}}
    \caption{Language Confusion Benchmark Line Pass Rate Reproduction for both Monolingual and Crosslingual settings.}
    \label{tab:lcb_lpr_reproduction}
\end{table*}

\begin{table}[!ht]
    \centering
     \resizebox{\linewidth}{!}{ 
     \begin{tabular}{l|>{\columncolor{gray!20}}ccccccccccccccccc}   
     \toprule
        \textbf{WPR}    & \textbf{AVG} & \textbf{Russian} & \textbf{Chinese} & \textbf{Korean} & \textbf{Arabic} & \textbf{Japanese} & \textbf{Hindi} \\ 
            \midrule
         \textit{\textbf{\underline{Monolingual}}} & ~ & ~ & ~ & ~ & ~ & ~ & ~ \\ 
        Command R & 96.31 & 96.00 & 92.50 & 97.00 & 99.33 & 94.00 & 99.00 \\ 
        Command R+ & 99.44 & 98.00 & 100.00 & 100.00 & 99.67 & 99.00 & 100.00 \\ 
        GPT-3.5 Turbo & 99.83 & 100.00 & 100.00 & 100.00 & 100.00 & 99.00 & 100.00 \\ 
        GPT-4 Turbo & 99.67 & 100.00 & 99.00 & 99.00 & 100.00 & 100.00 & 100.00 \\ 
        Mistral Large & 98.50 & 99.00 & 99.00 & 100.00 & 100.00 & 98.00 & 95.00 \\ 
        Mistral 8x7B & 73.75 & 83.00 & 64.50 & 62.00 & 86.00 & 68.00 & 79.00 \\ 
        Llama 2 70B-I & 97.92 & 93.00 & 94.50 & 100.00 & 100.00 & 100.00 & 100.00 \\ 
        Llama 3 70B-I & 93.19 & 94.00 & 93.50 & 100.00 & 95.67 & 80.00 & 96.00 \\ 
        \midrule
        \textit{\textbf{\underline{Crosslingual}}} & ~ & ~ & ~ & ~ & ~ & ~ & ~ \\ 
        Command R & 93.89 & 93.67 & 91.00 & 97.33 & 94.33 & 88.33 & 98.67 \\ 
        Command R+ & 95.11 & 89.67 & 95.67 & 96.00 & 98.00 & 95.33 & 96.00 \\ 
        GPT-3.5 Turbo & 98.72 & 98.33 & 99.00 & 98.33 & 99.00 & 98.67 & 99.00 \\ 
        GPT-4 Turbo & 96.61 & 95.67 & 97.00 & 96.67 & 97.33 & 95.67 & 97.33 \\ 
        Mistral Large & 93.83 & 93.00 & 97.67 & 92.00 & 93.67 & 91.33 & 95.33 \\ 
        Mistral 8x7B & 68.22 & 80.33 & 62.00 & 67.33 & 76.33 & 51.67 & 71.67 \\ 
        Llama 2 70B-I & 84.17 & 81.33 & 76.00 & 86.67 & 91.33 & 84.33 & 85.33 \\ 
        Llama 3 70B-I & 94.39 & 89.00 & 92.33 & 100.00 & 95.67 & 91.67 & 97.67 \\ 
        \bottomrule
    \end{tabular}
    }
    \caption{Language Confusion Benchmark Word Pass Rate Reproduction for both Monolingual and Crosslingual settings.}
    \label{tab:lcb_wpr_reproduction}
\end{table}


\begin{table}[!ht]
    \centering
     \resizebox{\linewidth}{!}{ 
    \begin{tabular}{c|l|c|c|c|c|c}
    \toprule
    
     \multirow{2}{*}{\textbf{Language}}& \multirow{2}{*}{\textbf{Step}}   & \multicolumn{2}{|c|}{\textbf{Monolingual}} & \multicolumn{2}{|c|}{\textbf{\underline{Crosslingual}}} &  \multirow{2}{*}{\textbf{BLEU (AVG)}} \\
         &  & \textbf{Line Level} & \textbf{Word Level} & \textbf{Line Level} & \textbf{Word Level} &\\ 
        \midrule 
        
        \multirow{3}{*}{\textbf{Sinhala}} & Base & - & - & \textbf{0.9061} & 1.6668 & 6.6147 \\ 
        & Step1 & - & - & 1.1412 & 1.7610 & 6.1822 \\ 
        & Step50+sbeam8 & - & - & 1.3213 & 1.9353 & 6.1517 \\ 
        \midrule
        
        \multirow{3}{*}{\textbf{Gujarati}}  & Base & 0.0385 & 0.2332 & \textbf{0.9372} & 1.7629 & 8.8752 \\ 
         & Step1 & 0.0308 & 0.9894 & 1.4092 & 1.7977 & 8.7153 \\ 
         & Step50+sbeam8 & 0.0154 & 0.2303 & 1.3821 & 1.5844 & 10.0989 \\ 
        \midrule
        
        \multirow{3}{*}{\textbf{Meadow Mari}} & Base & - & - & 1.1895 & 2.5281 & 6.8674 \\ 
        & Step1 & - & - & \textbf{1.3452} & 2.4527 & 6.3181 \\ 
         & Step50+sbeam8 & - & - & 1.4333 & 2.4806 & 6.0744 \\ 
         \midrule 
         
        \multirow{3}{*}{\textbf{Punjabi}}  & Base & 0 & 0.2304 & 1.2886 & 1.9215 & 11.7362 \\ 
         & Step1 & 0 & 0.1910 & 1.0879 & 1.7473 & 9.7027 \\ 
         & Step50+sbeam8 & 0 & 0.1994 & 0.8049 & 1.3955 & 11.4941 \\ 
         \midrule
         
        \multirow{3}{*}{\textbf{Amharic}}  & Base & - & - & 1.2949 & 1.9906 & 7.2613 \\ 
         & Step1 & - & - & 1.1420 & 1.7983 & 6.7393 \\ 
         & Step50+sbeam8 & - & - & 1.2548 & 1.8958 & 6.5141 \\ 
         \midrule
      \multirow{3}{*}{\textbf{Urdu}} & Base & 0.0516 & 0.7021 & 1.3266 & 2.3594 & 8.6339 \\ 
         & Step1 & 0.0462 & 0.5510 & 1.4870 & 2.8171 & 8.2143 \\ 
         & Step50+sbeam8 & 0.0385 & 0.5661 & 1.5844 & 2.3360 & 8.9373 \\ 
         \midrule
        \multirow{3}{*}{\textbf{Korean}} & Base & - & - & 1.4917 & 2.5477 & 5.5578 \\ 
         & Step1 & - & - & 1.6311 & 2.2823 & 5.1378 \\ 
         & Step50+sbeam8 & - & - & 1.6845 & 2.4643 & 4.9561 \\ 
         \midrule
       \multirow{3}{*}{\textbf{Yiddish}} & Base & - & - & 1.5216 & 2.3321 & 6.5486 \\ 
         & Step1 & - & - & 1.6912 & 2.2014 & 6.0670 \\ 
         & Step50+sbeam8 & - & - & 1.5956 & 2.2891 & 5.9480 \\ 
         \midrule
        \multirow{3}{*}{\textbf{Mongolian}}  & Base & 0.0462 & 2.6470 & 1.6223 & 2.6724 & 7.9523 \\ 
         & Step1 & 0.0462 & 2.6291 & 1.3172 & 2.5401 & 7.7916 \\ 
         & Step50+sbeam8 & 0.0462 & 2.6586 & 1.4105 & 2.6977 & 8.4952 \\ 

         \midrule
        \multirow{3}{*}{\textbf{Hindi}}& Base & 0.6587 & 1.0295 & 2.0050 & 3.0441 & 8.5273 \\ 
         & Step1 & 0.6305 & 1.0192 & 1.9432 & 2.5054 & 8.5912 \\ 
         & Step50+sbeam8 & 0.6212 & 1.0268 & 1.4697 & 2.6635 & 9.6811 \\ 
         \midrule
       \multirow{3}{*}{\textbf{Arabic}}& Base & 0.0280 & 0.5021 & 2.1931 & 3.0200 & 6.1455 \\ 
         & Step1 & 0 & 0.4597 & 1.7083 & 2.3915 & 5.9620 \\ 
         & Step50+sbeam8 & 0 & 0.4666 & 1.6406 & 2.1459 & 6.1903 \\ 
         \midrule
       \multirow{3}{*}{\textbf{Kazakh}}  & Base & 0.1390 & 1.5820 & 2.2796 & 3.1694 & 10.8145 \\ 
         & Step1 & 0.1070 & 1.4129 & 1.8063 & 3.0509 & 10.8421 \\ 
         & Step50+sbeam8 & 0.1098 & 1.4378 & 1.8065 & 3.0765 & 12.4864 \\ 
         \midrule
        \multirow{3}{*}{\textbf{Hebrew}}  & Base & 0 & 0.1555 & 2.8174 & 3.7918 & 5.9312 \\ 
         & Step1 & 0 & 0.1625 & 2.3715 & 3.3317 & 5.6492 \\ 
         & Step50+sbeam8 & 0 & 0.1391 & 2.0314 & 2.5126 & 5.7413 \\ 
                  \midrule

          \multirow{3}{*}{\textbf{Chinese}}  & Base & 0 & 0.6124 & 2.8718 & 3.2475 & 6.4620 \\ 
         & Step1 & 0.0280 & 0.5898 & 1.9708 & 2.3969 & 5.6923 \\ 
         & Step50+sbeam8 & 0.0280 & 0.6262 & 1.9363 & 2.2672 & 5.3919 \\ 
                  \midrule

          \multirow{3}{*}{\textbf{Maltese}}  & Base & - & - & 3.0059 & 4.1352 & 6.7777 \\ 
         & Step1 & - & - & 3.5456 & 4.5662 & 6.1049 \\ 
         & Step50+sbeam8 &  -&  - & 3.3048 & 4.5243 & 5.8991 \\ 
                  \midrule
                  
       \multirow{3}{*}{\textbf{Japanese}} & Base & 0.0693 & 0.8806 & 3.2545 & 4.9242 & 5.7113 \\ 
         & Step1 & 0.0693 & 0.8135 & 2.6345 & 3.2156 & 5.2014 \\ 
         & Step50+sbeam8 & 0.0693 & 0.8277 & 2.3426 & 3.1882 & 5.0411 \\ 
                  \midrule

        \multirow{3}{*}{\textbf{German}} & Base & 0.0000 & 0.6205 & 3.2764 & 4.6278 & 6.3678 \\ 
         & Step1 & 0.0231 & 0.6106 & 3.8041 & 4.4014 & 6.3865 \\ 
         & Step50+sbeam8 & 0 & 0.6203 & 3.3912 & 4.4915 & 6.8334 \\ 
                  \midrule

        \multirow{3}{*}{\textbf{Turkish}} & Base & 0.0660 & 1.6099 & 3.8939 & 4.8367 & 7.9789 \\ 
         & Step1 & 0.0707 & 1.6103 & 2.7842 & 3.7629 & 8.3830 \\ 
         & Step50+sbeam8 & 0.0687 & 1.6045 & 2.8521 & 4.0266 & 9.6997 \\ 
                  \midrule

       \multirow{3}{*}{ \textbf{Finnish}} & Base & - & - & 4.3696 & 5.6543 & 5.2877 \\ 
         & Step1 & - & - & 3.6533 & 5.0938 & 4.8876 \\ 
         & Step50+sbeam8 & - & - & 3.6939 & 5.0801 & 4.7593 \\ 
                  \midrule

       \multirow{3}{*}{ \textbf{Hungarian}} & Base & - & - & 4.4119 & 5.1123 & 5.1684 \\ 
         & Step1 & - & - & 3.6595 & 4.6260 & 4.7418 \\ 
         & Step50+sbeam8 & - & - & 3.3812 & 4.0739 & 4.6179 \\ 
        \bottomrule
    \end{tabular}
    }
    \caption{Language Confusion Entropy at each step in the Monolingual and Crosslingual generation settings at both Line and Word level, with BLEU score for textual embedding inversion tasks for \textbf{eval} languages.}
    \label{tab:mtei_lc_inversion_eval}
\end{table}

\begin{table}[!ht]
    \centering
         \resizebox{\linewidth}{!}{ 
    \begin{tabular}{c|c|l|c|c|c|c|c}
    \toprule
    \multirow{2}{*}{ \textbf{Train}} & \multirow{2}{*}{\textbf{Language}}& \multirow{2}{*}{\textbf{Step}}   & \multicolumn{2}{|c|}{\textbf{Monolingual}} & \multicolumn{2}{|c|}{\textbf{Crosslingual}} &  \multirow{2}{*}{\textbf{BLEU (AVG)}} \\
       &  &  & \textbf{Line Level} & \textbf{Word Level} & \textbf{Line Level} & \textbf{Word Level} &\\ 
      
        \midrule
        
         \multirow{36}{*}{\rotatebox{90}{  \textbf{Monolingual}}} & 
        \textbf{Punjabi} & Base & 0.0000 & 0.2615 & 0.3042 & 0.7821 & 6.7660 \\ 
       & & Step1 & 0.0000 & 0.0000 & 0.0000 & 0.0000 & 0.0227 \\ 
        && Step50+sbeam8 & 0.0000 & 0.0000 & 0.0000 & 0.0000 & 0.0229 \\ 
                \cmidrule(lr){2-8}

        &\textbf{Hebrew} & Base & 0.0000 & 0.1720 & 0.3404 & 0.8445 & 6.0112 \\ 
        & & Step1 & 0.0000 & 0.1391 & 0.4184 & 0.8806 & 5.9759 \\ 
        & & Step50+sbeam8 & 0.0000 & 0.1391 & 0.5401 & 0.9759 & 5.9736 \\
                \cmidrule(lr){2-8}

      &  \textbf{Gujarati} & Base & 0.0462 & 0.2332 & 0.7272 & 1.6964 & 6.3759 \\ 
       & & Step1 & 0.0462 & 0.2384 & 0.6853 & 1.3227 & 6.4278 \\ 
       & & Step50+sbeam8 & 0.0462 & 0.2384 & 0.4426 & 1.0568 & 6.4171 \\ 
               \cmidrule(lr){2-8}

       & \textbf{Chinese} & Base & 0.0000 & 0.5667 & 0.7711 & 1.3158 & 6.3859 \\ 
       & & Step1 & 0.0000 & 0.5667 & 0.6766 & 1.5313 & 6.1478 \\ 
       & & Step50+sbeam8 & 0.0560 & 0.6395 & 0.8930 & 1.2632 & 5.9179 \\ 
               \cmidrule(lr){2-8}

        &\textbf{German} & Base & 0.0000 & 0.6205 & 0.8030 & 1.4859 & 6.5072 \\ 
       & & Step1 & 0.0000 & 0.6008 & 0.9991 & 1.4383 & 6.4497 \\ 
       & & Step50+sbeam8 & 0.0000 & 0.6090 & 1.0879 & 1.4596 & 6.4801 \\ 
             \cmidrule(lr){2-8}

       & \textbf{Kazakh} & Base & 0.1252 & 0.9288 & 1.2115 & 2.1219 & 7.5891 \\ 
       & & Step1 & 0.1061 & 0.9288 & 0.7061 & 1.7686 & 7.5050 \\ 
       & & Step50+sbeam8 & 0.1252 & 0.9288 & 0.7562 & 1.7288 & 8.0075 \\
               \cmidrule(lr){2-8}

      &  \textbf{Urdu} & Base & 0.0462 & 0.7023 & 1.2381 & 1.9326 & 6.0338 \\ 
      &  & Step1 & 0.0462 & 0.6597 & 0.9932 & 1.8497 & 5.9788 \\ 
      &  & Step50+sbeam8 & 0.0462 & 0.6781 & 1.0471 & 1.8990 & 5.8903 \\ 
                \cmidrule(lr){2-8}

      &  \textbf{Japanese} & Base & 0.0925 & 0.8399 & 1.2677 & 2.5338 & 5.8317 \\ 
      &  & Step1 & 0.0925 & 0.8265 & 1.3099 & 2.3773 & 5.6813 \\ 
     &   & Step50+sbeam8 & 0.0925 & 0.8112 & 1.4349 & 2.5009 & 5.4885 \\ 
\cmidrule(lr){2-8}
       & \textbf{Turkish} & Base & 0.0925 & 1.0902 & 1.2699 & 2.1591 & 7.1899 \\ 
      &  & Step1 & 0.0462 & 1.0711 & 0.9757 & 1.8074 & 7.2374 \\ 
     &   & Step50+sbeam8 & 0.0462 & 1.0711 & 0.9824 & 2.0698 & 7.5047 \\ 
               \cmidrule(lr){2-8}

     &   \textbf{Hindi} & Base & 0.6952 & 1.0268 & 1.7053 & 2.0748 & 6.4369 \\ 
      &  & Step1 & 0.6276 & 0.9830 & 1.5004 & 1.8830 & 6.3956 \\ 
      &  & Step50+sbeam8 & 0.6417 & 1.0094 & 1.5470 & 1.9180 & 6.3725 \\ 
               \cmidrule(lr){2-8}

       & \textbf{Arabic} & Base & 0.0560 & 0.5021 & 1.8446 & 2.1560 & 6.0938 \\ 
      &  & Step1 & 0.0000 & 0.4471 & 0.8492 & 1.6087 & 6.0149 \\ 
      &  & Step50+sbeam8 & 0.0000 & 0.4610 & 1.1300 & 1.3489 & 6.1260 \\
               \cmidrule(lr){2-8}

      &  \textbf{Mongolian} & Base & 0.0462 & 0.7688 & 2.2537 & 2.9952 & 7.0101 \\ 
      &  & Step1 & 0.0462 & 0.7330 & 1.8297 & 2.9202 & 7.3608 \\ 
      &  & Step50+sbeam8 & 0.0462 & 0.7697 & 1.6677 & 2.5683 & 7.8355 \\ 
                \midrule

        \multirow{30}{*}{\rotatebox{90}{  \textbf{In-Family/In-Script}}} & 
        
        \textbf{Hindi, Punjabi} & Base & 0.3333 & 0.6300 & 2.6597 & 3.4826 & 8.1018 \\ 
       & & Step1 & 0.3317 & 0.6300 & 3.3548 & 3.7614 & 8.5398 \\ 
       & & Step50+sbeam8 & 0.3209 & 0.6300 & 3.4952 & 4.2414 & 9.2197 \\ 
                \cmidrule(lr){2-8}

       & \textbf{Punjabi, Urdu} & Base & 0.0393 & 0.4470 & 2.9478 & 3.9370 & 8.0232 \\ 
       & & Step1 & 0.0231 & 0.4361 & 4.0750 & 4.8806 & 7.3287 \\ 
       & & Step50+sbeam8 & 0.0231 & 0.4546 & 3.7656 & 4.6257 & 8.0605 \\ 
                \cmidrule(lr){2-8}

      &  \textbf{Kazakh, Turkish} & Base & 0.0995 & 1.0037 & 3.0674 & 5.2358 & 8.3012 \\ 
      &  & Step1 & 0.1158 & 0.9838 & 2.3818 & 4.7775 & 8.7420 \\ 
      &  & Step50+sbeam8 & 0.0924 & 1.0029 & 2.6656 & 4.2396 & 9.8136 \\ 
                \cmidrule(lr){2-8}

      &  \textbf{Gujarati, Hindi} & Base & 0.3209 & 0.6300 & 3.1485 & 4.7015 & 7.3518 \\ 
        
      &  & Step1 & 0.3332 & 0.6389 & 3.2087 & 3.5149 & 7.5794 \\ 
      &  & Step50+sbeam8 & 0.3369 & 0.6257 & 2.8926 & 3.1977 & 8.1357 \\ 
                 \cmidrule(lr){2-8}

      &  \textbf{German, Turkish} & Base & 0.0231 & 2.7236 & 3.3850 & 5.2398 & 7.5745 \\ 
       & & Step1 & 0.0693 & 2.7123 & 3.3267 & 5.2520 & 7.9995 \\ 
      &  & Step50+sbeam8 & 0.0393 & 2.7291 & 3.0071 & 5.3435 & 8.9119 \\
       \cmidrule(lr){2-8}
       
      &  \textbf{Gujarati, Urdu} & Base & 0.0462 & 0.4768 & 3.6012 & 4.2957 & 7.5971 \\ 
      &  & Step1 & 0.0231 & 0.4678 & 3.6518 & 4.7169 & 7.8532 \\ 
      &  & Step50+sbeam8 & 0.0231 & 0.4579 & 3.2620 & 4.0107 & 8.3644 \\ 
        \cmidrule(lr){2-8}
        
      &  \textbf{Hindi, Urdu} & Base & 0.3327 & 0.8645 & 3.6261 & 4.0579 & 7.4791 \\ 
      &  & Step1 & 0.3258 & 0.8347 & 3.5240 & 3.8894 & 7.7025 \\ 
      &  & Step50+sbeam8 & 0.3064 & 0.8460 & 3.2558 & 3.7994 & 8.1493 \\
         \cmidrule(lr){2-8}
        
      &  \textbf{Arabic, Hebrew} & Base & 0.0000 & 0.3206 & 3.9135 & 4.6422 & 7.2555 \\ 
      &  & Step1 & 0.0000 & 0.3290 & 3.0216 & 3.8845 & 7.1863 \\ 
      &  & Step50+sbeam8 & 0.0000 & 0.3056 & 3.0272 & 4.1192 & 7.3948 \\ 
                \cmidrule(lr){2-8}

      &  \textbf{Gujarati, Punjabi} & Base & 0.0231 & 0.2332 & 4.3682 & 5.0570 & 7.9168 \\ 
      &  & Step1 & 0.0231 & 2.4993 & 3.1955 & 4.0486 & 8.2486 \\ 
      &  & Step50+sbeam8 & 0.0000 & 0.2220 & 2.7857 & 4.2355 & 9.1628 \\ 
    \cmidrule(lr){2-8}

      &  \textbf{Kazakh, Mongolian} & Base & 0.0995 & 4.9836 & 4.6250 & 5.5072 & 9.0396 \\ 
      &  & Step1 & 0.0761 & 4.9836 & 4.3483 & 5.1664 & 9.6267 \\ 
      &  & Step50+sbeam8 & 0.0995 & 4.9947 & 3.6838 & 5.6001 & 10.8565 \\ 
                \midrule
                
      \multirow{27}{*}{\rotatebox{90}{  \textbf{Cross-Family/Cross-Script}}} & \textbf{Gujarati, Turkish} & Base & 0.0693 & 0.6534 & 1.7507 & 3.1341  & 7.3780 \\ 
      &   & Step1 & 0.0463 & 0.6525 & 2.4953 & 3.1636 & 7.6174 \\ 
      &   & Step50+sbeam8 & 0.0463 & 0.6409 & 2.1876 & 3.1276 & 8.1549 \\ 
                 \cmidrule(lr){2-8}

      &   \textbf{Chinese, Japanese} & Base & 0.0231 & 0.7897 & 1.9550 & 3.2433 & 6.4919 \\ 
     &    & Step1 & 0.0511 & 0.7067 & 3.2448 & 4.7459 & 6.5438 \\ 
     &    & Step50+sbeam8 & 0.0231 & 0.7285 & 3.7689 & 4.0670 & 6.6168 \\ 
                \cmidrule(lr){2-8}

      &   \textbf{Punjabi, Turkish} & Base & 0.0231 & 0.6417 & 2.1019 & 3.4377 & 8.0628 \\ 
     &    & Step1 & 0.0231 & 0.6440 & 2.8690 & 3.6396 & 7.2908 \\ 
     &    & Step50+sbeam8 & 0.0231 & 0.6548 & 2.5556 & 3.6179 & 7.9988 \\ 
                \cmidrule(lr){2-8}

     &    \textbf{Hindi, Turkish} & Base & 0.3682 & 1.0384 & 2.3767 & 3.5239 & - \\ 
     &    & Step1 & 0.3489 & 1.0369 & 2.2821 & 3.1379 & - \\ 
    &     & Step50+sbeam8 & 0.3601 & 1.0321 & 2.3311 & 3.1449 & - \\ 
                \cmidrule(lr){2-8}

      &   \textbf{Hindi, Kazakh} & Base & 0.4053 & 0.9859 & 2.8022 & 4.7000 & - \\ 
      &   & Step1 & 0.4069 & 0.9749 & 3.2070 & 4.3596 &  - \\ 
     &    & Step50+sbeam8 & 0.3735 & 0.9994 & 3.4030 & 3.9206 & - \\ 
                \cmidrule(lr){2-8}

      &   \textbf{Gujarati, Kazakh} & Base & 0.0761 & 0.5911 & 2.8922 & 3.9972 & 7.8211 \\ 
     &    & Step1 & 0.0857 & 0.5548 & 2.4678 & 3.0565 & 8.2088 \\ 
    &     & Step50+sbeam8 & 0.0531 & 0.5810 & 2.0942 & 2.9441 & 8.8596 \\ 
        \cmidrule(lr){2-8}
        
     &    \textbf{Turkish, Urdu} & Base & 0.0462 & 0.9011 & 2.9759 & 4.0922 & 7.5098 \\ 
    &     & Step1 & 0.0693 & 0.8723 & 2.3225 & 3.3931 & 7.7308 \\ 
    &     & Step50+sbeam8 & 0.0462 & 0.9011 & 2.1042 & 3.3795 & 8.1401 \\ 
        \cmidrule(lr){2-8}
   &     \textbf{Kazakh, Punjabi} & Base & 0.0764 & 0.5697 & 3.2837 & 4.0113 & 8.4029 \\ 
    &     & Step1 & 0.0531 & 0.5435 & 2.2224 & 3.2103 & 8.6699 \\ 
    &     & Step50+sbeam8 & 0.0764 & 0.5705 & 1.6113 & 3.1262 & 9.3574 \\ 
      
        \bottomrule
    \end{tabular}
    }
     \caption{Language Confusion Entropy at each step in the Monolingual and Crosslingual generation settings at both Line and Word level, with BLEU score for textual embedding inversion tasks for \textbf{train} languages.}
     \label{tab:mtei_lc_bleu_train}
\end{table}

    


\section{Language Graphs for Language Similarities}~\label{sec:language_graphs}
We curate language graphs from a diverse range of sources, as shown in Table~\ref{tab:language_graph_stats}.
The language vectors from Grambank and WALS consist of multi-valued features, while those derived from colexification patterns in CLICS$^3$ and WordNet (WN) are binarized. For these, we employ the Jaccard index to compute pairwise language similarities.
For other more dense valued language vectors, we use cosine similarity instead.




\begin{table*}[ht!]
    \centering
   \resizebox{\linewidth}{!}{ 
    \begin{tabular}{l|ccp{3cm}p{5cm}p{6cm}c}
    \toprule
       \textbf{Language Graph}  &  \textbf{\#Languages} & \textbf{\#Features} & \textbf{Category of Features} & \textbf{Datasource References} & \textbf{Representation} & \textbf{Similarity Metric}\\
       \midrule
        \textbf{Grambank} & 2292 & 195 & Morpho-syntactical &\citet{skirgaard2023grambank} & Multivalued Vectors  & Jaccard \\
           \textbf{WALS} & 424 & 192 & Structural properties of languages &~\citet{haspelmath2008typological}& Multivalued Vectors &  Jaccard\\
            \textbf{WALS$\backslash$ Phon.} &  431 & 172 & WALS w/o Phonological Features  & ~\citet{haspelmath2008typological} & Multivalued Vectors & Jaccard \\
            
            \midrule 
            \underline{\textbf{Lang2Vec}}~\citep{littell2017uriel}& 8070 & - & Inventory, Syntax, Phonology, Genealogy, Geography, Featural & ~\citet{collin2010ethnologue, haspelmath2008typological,collins2011syntactic} & Vectors & Cosine similarity and Arccosine\\
            \midrule 
             \textbf{$\text{CLICS}^3$} & 1347 & 4228  & Colexifications &~\citet{rzymski2020database}  & Binarized Vectors  & Jaccard   \\
              \textbf{WN}&  519 & 2,518,357 & Colexifications & ~\citet{navigli2010babelnet} & Binarized Vectors& Jaccard \\
              \midrule 

               \underline{\textbf{Colex2Lang}}~\citep{chen-etal-2023-colex2lang}& \\
            \textbf{$\text{CLICS}^3$} &1609  & 4228 &Colexifications &~\citet{rzymski2020database} & Combined Graph Embeddings &  Cosine Similarity  \\
            
            \textbf{WN} & 519 & 2,525,591& Colexifications&  ~\citet{navigli2010babelnet}& Combined Graph Embeddings&  Cosine Similarity  \\
            \textbf{WN\_CONCEPT} & 519 &2,486,485 & Colexifications&  ~\citet{navigli2010babelnet}& Combined Graph Embeddings &  Cosine Similarity  \\
            \midrule
            \textbf{ASJP SVD} &  1012 &40 & lexicon &~\citep{wichmann2012automated} & UMAP on a mean normalized Levenshtein distance pairwise distance matrix from ASJP &  Cosine Similarity  \\
            \textbf{ASJP UMAP} & 1012  & 40 & lexicon &~\citep{wichmann2012automated}&Truncated SVD on a mean normalized Levenshtein distance pairwise distance matrix from word alignments  &  Cosine Similarity  \\

            \midrule
            \citet{ostling-tiedemann-2017-continuous} & 943 & - &-& Bible Translations & Lang. Vectors trained on Bible Data&  Cosine Similarity  \\            
            \bottomrule
    \end{tabular}
    }
    \caption{Statistics and Incorporated Features for Language Graphs.}
    \label{tab:language_graph_stats}
\end{table*}


\begin{table*}[!ht]
    \centering
      \resizebox{\linewidth}{!}{ 
    \begin{tabular}{l|>{\columncolor{gray!20}}c|cc|cc|cc||>{\columncolor{gray!20}}c|cc|cc|cc}
    \toprule
    
    \multirow{3}{*}{\textbf{Language Graph}}& \multicolumn{7}{|c||}{\textbf{LCB}}   & \multicolumn{7}{c}{\textbf{Textual Embedding Inversion}} \\
        &  \textbf{AVG} & \multicolumn{2}{|c|}{\textbf{ALL}} &\multicolumn{2}{|c|}{\textbf{Monolingual}} &\multicolumn{2}{|c||}{\textbf{Crosslingual}} & \textbf{AVG} &  \multicolumn{2}{c|}{\textbf{ALL}} &\multicolumn{2}{|c|}{\textbf{Monolingual}} &\multicolumn{2}{|c}{\textbf{Crosslingual}} \\
        
         & & Line & Word &  Line & Word  &  Line & Word  & & Line & Word  &  Line & Word  &  Line & Word \\ 
         \midrule
           
        \textbf{Grambank} &0.2650&  0.1830 & 0.2827 & 0.2726 & 0.4042 & 0.1791 & 0.2682 & 0.7538 & \textbf{0.8005} & 0.7891 & 0.7504 & 0.5434 & \textbf{0.8230} & 0.8162 \\ 
        \textbf{WALS} &0.1947 & 0.0420&	0.2908	& 0.0816 &0.4286&  0.0388 &	0.2865 & 0.7377 & 0.8722 & \textbf{0.7854} & 0.6618 & 0.4102 & \underline{0.8803} & 0.8164 \\ 
        
        \textbf{WALS $\backslash$ Phon.} & 0.1892&  0.0409 &	0.2889&	0.0799&	0.4253	& 0.0379 &	0.2620 & \underline{0.7360} & 0.8768 & \underline{0.7880} & 0.6495 & 0.4036 & 0.8837 & \underline{0.8147} \\

        \midrule
    \underline{\textbf{Lang2Vec}} &  &   &    &   &  &   & &   & &  &   &  &  &   \\       
\textbf{Inventory}  & 0.1650 &0.0812 & \textbf{0.2003} & 0.1374 & \textbf{0.3114} & 0.0637 & \textbf{0.1961} &0.8154 & 0.9678 & 0.8410 & 0.8225 & 0.4124 & 0.9720 & 0.8768 \\ 
        \textbf{Syntactic} &0.2925&  0.1949 & 0.3771 & 0.2244 & 0.4679 & 0.1505 & 0.3405 & 0.8651& 1.0179 & 0.8668 & 0.8872 & 0.5447 & 1.0001 & 0.8742 \\ 
        \textbf{Phonological} &0.2260& 0.1009 & 0.3126 & 0.1552 & 0.4235 & 0.0830 & 0.2808 & 1.3161& 1.7178 & 1.6133 & 0.6859 & 0.6004 & 1.6599 & 1.6191 \\ 
        \textbf{Genetic} & 12.8307 & 13.4548 & 12.2278 & 12.9350 & 12.3563 & 13.5850 & 12.4249 & 14.9443& 14.6988 & 14.9756 & 18.5227 & 13.5734 & 13.8721 & 14.0232 \\ 
        \textbf{Geographical} & 1.5976 & 1.5924 & 1.8769 & 0.9753 & 1.8614 & 1.4996 & 1.7798 & 2.5218 & 2.4910 & 2.3964 & 2.6389 & 2.7875 & 2.4732 & 2.3439 \\ 
        \textbf{Featural} &0.3174&  0.2094 & 0.4099 & 0.2429 & 0.5033 & 0.1626 & 0.3762 & 1.0628 & 1.0940 & 0.9355 & 1.5467 & 0.7976 & 1.0673 & 0.9359 \\ 
        \midrule
      
        
         \textbf{$\text{CLICS}^3$}  &1.6777& 1.6516 & 1.7336 & 1.5453 & 1.7043 & 1.7182 & 1.7130 & 1.1806 & 1.1601 & 1.0850 & 1.4595 & 1.2880 & 1.0670 & 1.0237 \\ 
        \textbf{WN} &1.1372& 1.1211 & 1.1423 & 1.1491 & 1.2521 & 1.0906 & 1.0680 &2.1364 & 2.2197 & 2.0398 & 2.9402 & 1.6996 & 2.0687 & 1.8501 \\ 

        \midrule
     \underline{\textbf{Colex2Lang}} &  &   &    &   &  &   & &   & &  &   &  &  &   \\

        \textbf{$\text{CLICS}^3$} & 0.1665 & 0.0289 & 0.2522 & \underline{0.0589} & 0.3776 & 0.0260 & 0.2490 & \textbf{0.7333} & 0.9492 & 0.8335 & 0.3690 & \textbf{0.2672} & 0.9503 & 0.8693 \\ 
        \textbf{WN} & \textbf{0.1489} & \textbf{0.0242} & \underline{0.2154} & \textbf{0.0522} & \underline{0.3404} & \textbf{0.0218} & \underline{0.2149} & 0.7794 & 1.0105 & 0.8892 & 0.3263  & \underline{0.3894} & 1.0445 & 0.9454 \\ 
        \textbf{WN\_CONCEPT} & \underline{0.1490} & \textbf{0.0242} & 0.2175 & 0.0522 & 0.3426 & \textbf{0.0218} & 0.2168 & 0.7791 & 1.0100 & 0.8836 & \textbf{0.3263} & \underline{0.3894} & 1.0427 & 0.9390 \\ 
         \midrule
        \textbf{ASJP SVD} & 0.5859 & \underline{0.0253} & 1.0597 & \textbf{0.0522} & 1.0395 & \underline{0.0225} & 1.3164 & 2.8254 & 2.9730 & 3.3461 & 0.0820 & 1.6568 & 3.8882 & 5.0063 \\ 
        \textbf{ASJP UMAP} & 0.9318 & 0.0994 & 1.6767 & 0.1290 & 1.6055 & 0.0946 & 1.9856 & 3.3217 & 3.5364 & 4.0245 & 0.0811 & 2.3479 & 4.3690 & 5.5711 \\ 
        \midrule
        \multicolumn{3}{l}{\underline{\textbf{\citet{ostling-tiedemann-2017-continuous}}} }&     &      &  &   & &   & &  &   &  &  &   \\
        \textbf{L1} & 0.5697 & 0.0504 & 0.9843 & 0.0920 & 0.9616 & 0.0483 & 1.2819 & 4.7062 & 5.3973 & 5.9836 & 0.2837 & 1.7125 & 7.1117 & 7.7482 \\ 
        \textbf{L2} & 0.5644 & 0.0497 & 0.9749 & 0.0918 & 0.9536 & 0.0452 & 1.2710 & 4.6788 & 5.4149 & 5.9684 & \underline{0.1421} & 1.7019 & 7.1158 & 7.7296 \\ 
        \textbf{L3} & 0.5589 & 0.0534 & 0.9601 & 0.0981 & 0.9336 & 0.0523 & 1.2558 & 4.7099 & 5.4686 & 6.0242 & \textbf{0.1138} & 1.6944 & 7.1689 & 7.7895 \\ 
    
        \textbf{ALL} & 0.5595 & 0.0471 & 0.9673 & 0.0900 & 0.9437 & 0.0450 & 1.2641 & 4.6870 & 5.4175 & 5.9885 & 0.1482 & 1.6916 & 7.1240 & 7.7525 \\

        \bottomrule 
    \end{tabular}
    }
    \caption{KL Divergence between Language Similarity Graphs and the Language Confusion Matrices for Target/ Eval Languages from LCB and Inversion Tasks. The best results (lowest) are \textbf{bolded}, and the second best are \underline{underlined}.}
    \label{tab:kl_all}
\end{table*}
\end{document}